\documentclass[lettersize,journal]{IEEEtran}
\usepackage{amsmath,amsfonts}
\usepackage{algorithmic}
\usepackage{algorithm}
\usepackage{array}
\usepackage[caption=false,font=normalsize,labelfont=sf,textfont=sf]{subfig}
\usepackage{textcomp}
\usepackage{stfloats}
\usepackage{url}
\usepackage{verbatim}
\usepackage{graphicx}
\usepackage{cite}
 \usepackage{array}
\usepackage{arydshln}
\usepackage{newfloat}
\usepackage{listings}
\usepackage{multirow}
\usepackage{svg}
\usepackage{booktabs}

\begin{document}

\title{Optimizing Soft Prompt Tuning via Structural Evolution}

\author{Zhenzhen Huang, Chaoning Zhang*,~\IEEEmembership{Senior Member, IEEE,} Haoyu Bian, Songbo Zhang, Chi-lok Andy Tai, Jiaquan Zhang, Caiyan Qin, Jingjing Qu, Yalan Ye, Yang Yang,~\IEEEmembership{Senior Member, IEEE} and Heng Tao Shen,~\IEEEmembership{Fellow, IEEE}

\thanks{Zhenzhen Huang, Songbo Zhang, and Jiaquan Zhang are with School of Information and Software Engineering, University of Electronic Science and Technology of China, Chengdu, China (email: alley10086@gmail.com; 2024090914002@std.uestc.edu.cn; jiaquanzhang2005@gmail.com).
Chaoning Zhang, Haoyu Bian, Yalan Ye, and Yang Yang are with School of Computer Science and Engineering, University of Electronic Science and Technology of China, Chengdu, China (email: chaoningzhang@uestc.edu.cn; haoyubian04@gmail.com; yalanye@uestc.edu.cn; yang.yang@uestc.edu.cn).
Chi-lok Andy Tai is with College of Professional and Continuing Education, The Hong Kong Polytechnic University, Hong Kong, China (email: andy.tai@cpce-polyu.edu.hk).
Caiyan Qin is with School of Robotics and Advanced Manufacture, Harbin Institute of Technology, Shenzhen, China (email: qincaiyan@hit.edu.cn). 
Jingjing Qu is with Shanghai Artificial Intelligence Laboratory, Shanghai, China (e-mail:qujingjing@pjlab.org.cn).
Heng Tao Shen is with School of Computer Science and Technology, Tongji University, Shanghai, China (e-mail: shenhengtao@tongji.edu.cn).}
\thanks{* Corresponding Author}
}



\maketitle

\begin{abstract}
Soft prompt tuning leverages continuous embeddings to capture task-specific information in large pre-trained language models (LLMs), achieving competitive performance in few-shot settings. However, soft prompts rely on high-dimensional, implicit representations and lack explicit semantics and traceable training behaviors, which limits their interpretability. To address this limitation, we propose a soft prompt tuning optimization method based on topological morphological evolution. Specifically, we employ persistent homology from topological data analysis (TDA) to quantify the structural representations of soft prompts in continuous parameter space and their training process evolution. Quantitative analysis shows that topologically stable and compact soft prompts achieve better downstream performance. Based on this empirical observation, we construct a loss function for optimizing soft prompt tuning, termed Topological Soft Prompt Loss (TSLoss). TSLoss guides the model to learn structurally stable adaptations by quantifying inter-parameter connectivity and redundancy. Extensive experiments show that training with TSLoss accelerates convergence and improves tuning performance, providing an interpretable method to understand and optimize soft prompt tuning from structural and topological perspectives.
\end{abstract}


\section{Introduction}
\IEEEPARstart{P}{rompt} Tuning (PT) \cite{lester2021power} offers high parameter efficiency, which reduces computational and memory requirements and enables flexible deployment for multiple tasks. A common form of PT is soft prompt tuning \cite{vu2021spot,han2024parameter}, which operates in the model's embedding space and requires only a few trainable continuous prompt vectors added to the input to achieve effects comparable to full parameter fine-tuning. Existing research demonstrates their superiority through high-quality soft prompts and consistently improves task-specific accuracy across diverse benchmarks \cite{vu2021spot,bai2024soft}.


However, the design of soft prompts results in a lack of transparency and interpretability compared with manually crafted or discrete natural language prompts. Specifically, soft prompts consist of trainable continuous vectors residing in dense high-dimensional embedding spaces that are not directly interpretable by humans \cite{schulhoff2024prompt}. Moreover, since LLMs remain frozen during the prompt tuning process and only soft prompt embeddings are updated, it becomes difficult to determine the specific convergence trajectories and transformations of the vectors \cite{vu2022spot}. Consequently, existing methods rely primarily on accuracy-based metrics to assess whether soft prompts successfully guide LLMs' reasoning states. This degree of limited interpretability introduces potential risks for practical applications, particularly in safety-critical domains such as healthcare, finance, and public policy, where transparent decision-making is essential \cite{wang2023universality}.

Current research explores the interpretability of soft prompts to improve their trustworthiness, effectiveness, and generalization capabilities. PGSPT \cite{liou2025semantically} projects soft prompt vectors onto fixed word embedding spaces and uses cosine similarity to identify their semantically most similar tokens, bridging latent space and textual semantics. Patel et al. \cite{patel2025towards} introduced Hard Prompts Made Easy (PEZ) and RLPrompt, revealing the inherent connection between performance and interpretability in prompt tuning tasks. Dynamic Prompt Perturbation (DPC) \cite{fan2025improving} dynamically evaluate the positive and negative impacts of soft prompts during reasoning and selectively mask interfering information, thereby enhancing both the performance and interpretability of LLMs on reasoning tasks. Nevertheless, existing work does not directly focus on vector analysis in the high-dimensional semantic space of soft prompts, thereby failing to reveal structural differences during training or elucidate the mechanism underlying correct model reasoning.


To address the interpretability gap in soft prompt tuning, we quantitatively analyze the evolution of soft prompt representations in the continuous parameter space during training and construct a loss function for optimization. We treat a soft prompt as a set of trainable embedding vectors and model its training dynamics as a trajectory of point clouds in a high-dimensional space. Specifically, we introduce persistent homology from topological data analysis (TDA) \cite{zomorodian2004computing, chazal2021introduction} to characterize the multiscale connectivity and redundancy patterns of prompt vectors throughout training. 
Quantitative results reveal that effective tuning processes are consistently accompanied by the gradual emergence of structurally stable and compact representations, manifested by the persistence of essential low-dimensional topological features and the elimination of short-lived redundant structures.

Motivated by this observation, we propose the Topological Soft Prompt Loss (TSLoss), which explicitly regularizes prompt representations toward simplified yet well-connected structures by discouraging redundant structural patterns while preserving essential connectivity. In doing so, TSLoss aligns the optimization objective with structural properties commonly observed in effective soft prompt tuning. Extensive experimental results and theoretical analysis reveal a consistent relationship between the structural organization of prompt representations and tuning performance, and show that TSLoss improves both convergence efficiency and final accuracy. From a structural perspective, this work provides an interpretable framework for understanding and optimizing soft prompt tuning.  Our contributions are as follows:

\begin{itemize}
    \item We propose an interpretable structural analysis framework for soft prompt tuning based on persistent homology, revealing the relationship between representation structure and tuning performance.
    \item We propose TSLoss, a loss function for soft prompt tuning, which facilitates convergence and improves accuracy.
    \item Extensive experiments and theoretical analysis confirm the validity of the quantified structural properties and the effectiveness of TSLoss.
\end{itemize}


\section{Related Work}
\label{sec:2}
\subsection{Soft Prompt Tuning}
Soft prompts progress from static to dynamic approaches, improving computational efficiency and task adaptability \cite{mangrulkar2022peft}. Early work focuses on static vector optimization, which aims to adapt pre-trained language models using fixed input embeddings, while Prompt Tuning \cite{lester2021power} adds trainable vectors to the input, and Prefix-Tuning \cite{li2021prefix} extends trainable parameters across attention layers. Despite these advances, Prompt Tuning and Prefix-Tuning continue to face challenges in cross-task generalization and interpretability. The latest research directions include cross-modal extensions and automated prompt design. LASP \cite{bulat2022language} constrains visual-language prompt semantics using a text alignment loss, and the Automated Prompt Optimization Framework \cite{murthy2025promptomatix} enables zero-configuration prompt engineering, reducing the need for manual intervention.

Soft prompt tuning exhibits effectiveness and generalization capabilities on downstream tasks. Existing applications focus on extending soft prompts to generative tasks, including question answering, summarization, and conditional text generation \cite{chang2024efficient}. In low-resource and domain transfer scenarios, soft prompt tuning exhibits adaptability \cite{philippy2025enhancing,zhu2024iterative}. By compressing task-related information into prompt vectors, soft prompt tuning achieves rapid transfer with limited labeled data and supports efficient reuse across tasks and domains \cite{vykopal2025soft}. Additionally, in multilingual and cross-lingual tasks, soft prompts effectively learn task representations with strong transferability, leading to consistently improved generalization compared to discrete prompts \cite{li2023enhancing}.

Recently, soft prompt methods have expanded to multimodal and reasoning scenarios. Vision-language soft prompts align multimodal representation spaces through continuous prompts to guide task execution in multimodal models \cite{chen2025trisprompt}; dynamic or interventional prompt methods regulate prompt representations during reasoning to enhance model reasoning stability and robustness \cite{yang2024soft}. Overall, existing research validates the practical value of soft prompt tuning across different tasks, data scales, and modal settings, establishing it as a parameter-efficient interface connecting pre-trained models with downstream tasks.

Various approaches have emerged to study and optimize the internal mechanisms of soft prompts. Dynamic intervention techniques like DPC \cite{fan2025improving} optimize LLMs' reasoning capabilities by selectively suppressing redundant information flow in the embedding space. Recent work has addressed soft prompt interpretability through theoretical frameworks that reveal trade-offs between interpretability and performance \cite{patel2025towards}. Geometric approaches to prompting have further uncovered distinct representational mechanisms for task adaptation, highlighting how different prompting methods affect representation geometry and the critical role of input distribution samples in few-shot learning contexts\cite{kirsanov2025geometry}. However, existing research lacks a comprehensive internal analysis of soft prompts, particularly from a shape characteristic perspective, largely due to the inherent invisibility of these high-dimensional representations.
Various approaches have emerged to study and optimize the internal mechanisms of soft prompts. Dynamic intervention techniques such as DPC \cite{fan2025improving} improve LLMs' reasoning capabilities by selectively suppressing redundant information flow in the embedding space. Recent studies investigate soft prompt interpretability through theoretical frameworks that reveal trade-offs between interpretability and performance \cite{patel2025towards}. Geometry-based approaches investigate how prompting strategies affect representation geometry and task adaptation, highlighting the role of input distribution in few-shot learning \cite{kirsanov2025geometry}. Despite these advances, a comprehensive internal analysis of soft prompts, especially from a shape perspective, remains lacking due to the inherent invisibility of high-dimensional embeddings.

\subsection{TDA for Natural Language Process (NLP) Analysis}
Topological Data Analysis (TDA) offers a framework for understanding complex linguistic structures by quantifying shape and connectivity patterns in language data. This method identifies global structural patterns that are invariant under transformations and responsive to relevant structural changes. Representative TDA algorithms include persistent homology \cite{zomorodian2004computing} for capturing multi-scale connectivity and cyclic structures, the Mapper algorithm \cite{madukpe2026comprehensive} for high-dimensional point cloud visualization, and Vietoris-Rips complexes \cite{zomorodian2010fast} for constructing simplicial complexes to analyze topological features. 

\begin{figure*}[h]
  \centering
  \includegraphics[width=\linewidth]{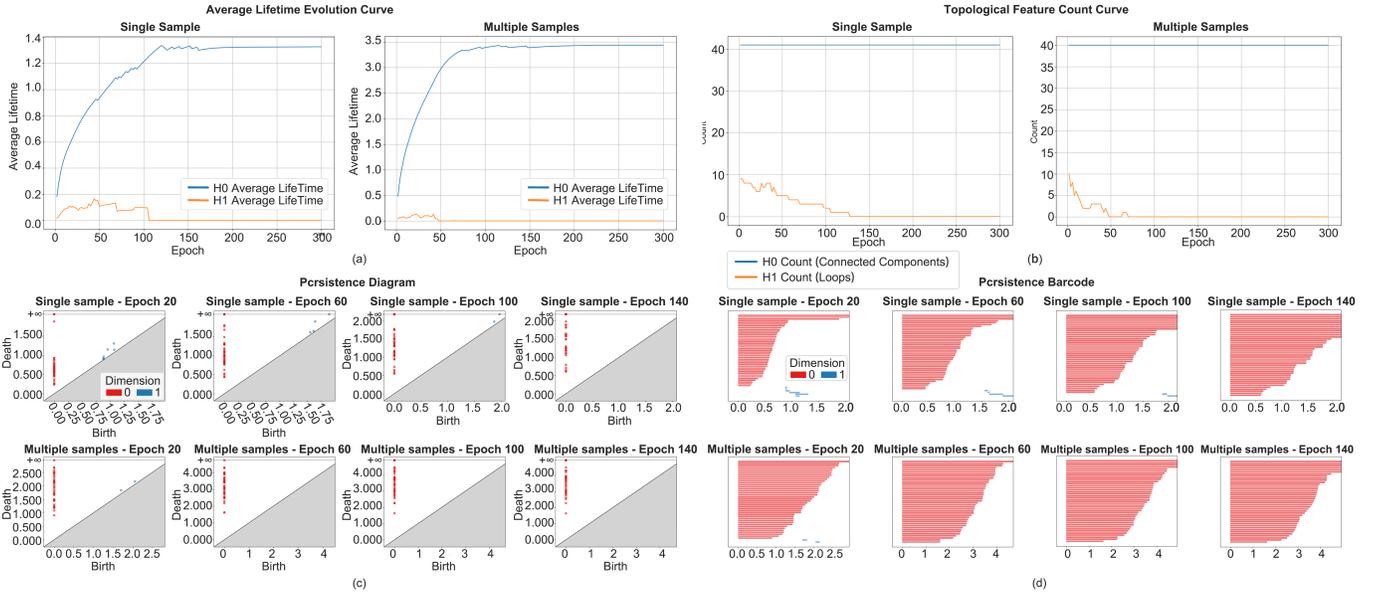}
  \caption{Topological evolution of soft prompts parameter space during training process.}
  \label{fig:1}
\end{figure*}

Unlike traditional methods that primarily rely on statistical distributions or contextual similarities, TDA analyzes multiscale topological features using tools such as persistent homology and the Mapper algorithm, uncovering structural information missed by conventional approaches \cite{michel2017does}. Recent applications demonstrate TDA's effectiveness across diverse NLP tasks, with researchers leveraging topological signatures to enhance sentiment analysis performance by capturing emotional trajectories in text \cite{gholizadeh2020novel}, detect AI-generated content through topological inconsistencies in embedding spaces \cite{uchendu2023topformer}, and interpret attention mechanisms in transformer architectures \cite{kushnareva2021artificial}. The topology-preserving properties of TDA make it particularly valuable for analyzing cross-lingual phenomena, as shown by Port et al. \cite{port2022topological} who identified invariant syntactic structures across language families that persist despite surface differences. As LLMs continue to scale, TDA provides a framework to study emergent properties by analyzing the topological evolution of their representation spaces. This includes scalable pipelines that characterize the birth and death of features across transformer layers \cite{gardinazzi2024persistent}.

\section{Structural Analysis of Soft Prompt Tuning}
\label{sec:3}
In this section, we analyze the evolution of soft prompt representations during the training process and extract structural regularities that correlate with tuning performance.

\subsection{Structural Quantification}

Soft prompts are neural network-based prompts that insert specific vectors into the input sequence to guide the model's attention to key information. This approach improves model performance and enhances adaptability to new tasks through fine-tuning. In the parameter space of soft prompt tuning, a prompt is represented as a learnable parameter matrix \( P \in \mathbb{R}^{n \times d} \), where \( n \) denotes the number of tokens and \( d \) is the model's hidden dimension. Each row of the matrix is interpreted as a point in a \( d \)-dimensional embedding space. Therefore, a soft prompt corresponds to \( n \) points, and the topological relationships among these points capture the relative relations and distribution patterns of the prompt vectors.

Topological data analysis (TDA) provides tools to quantify these structural features across multiple scales. We employ persistent homology to track the lifecycle of topological features from formation to disappearance, identify stable structural patterns, and filter transient noise. Specifically, Vietoris-Rips complexes are constructed with increasing distance thresholds \( \epsilon \) to form a filtration sequence that captures multiscale proximity relations among vectors. Homology groups are computed at each scale, where \( H_0 \) represents connected components and indicates the stability of soft prompts, and \( H_1 \) represents loops or redundant structures corresponding to structural redundancy. Persistence measures the lifespan of each feature to quantify structural stability and complexity. Metrics such as persistence entropy and average lifespan provide a quantitative analysis of the topological patterns during training and identify stable structures associated with model performance. The notation of the all metrics is summarized in Table \ref{tab:notation}.

\begin{table}[h!]
\centering
\caption{Notation summary for topological analysis of soft prompts}
\label{tab:notation}
\begin{tabular}{ll}
\hline
\multicolumn{2}{c}{\textbf{Topological Data Analysis Notation}} \\
\hline
Symbol & Description \\
\hline
$H_0$ & Zero-dimensional homology group \\
$H_1$ & First-dimensional homology group  \\
$(b_i, d_i)$ & Birth-death pair of $i$-th topological feature \\
$l_i$ & Lifespan of topological feature ($l_i = d_i - b_i$) \\
$L$ & Total lifespan ($L = \sum_i l_i$) \\
PE & Persistence entropy ($-\sum_i \frac{l_i}{L} \log \frac{l_i}{L}$) \\
$\epsilon$ & Filtration parameter (neighborhood radius) \\
$\partial_k$ & Boundary operator of dimension $k$ \\
\hline
\end{tabular}
\end{table}

\subsection{Empirical Observations}
\begin{figure}[h!]
  \centering
  \includegraphics[width=\linewidth]{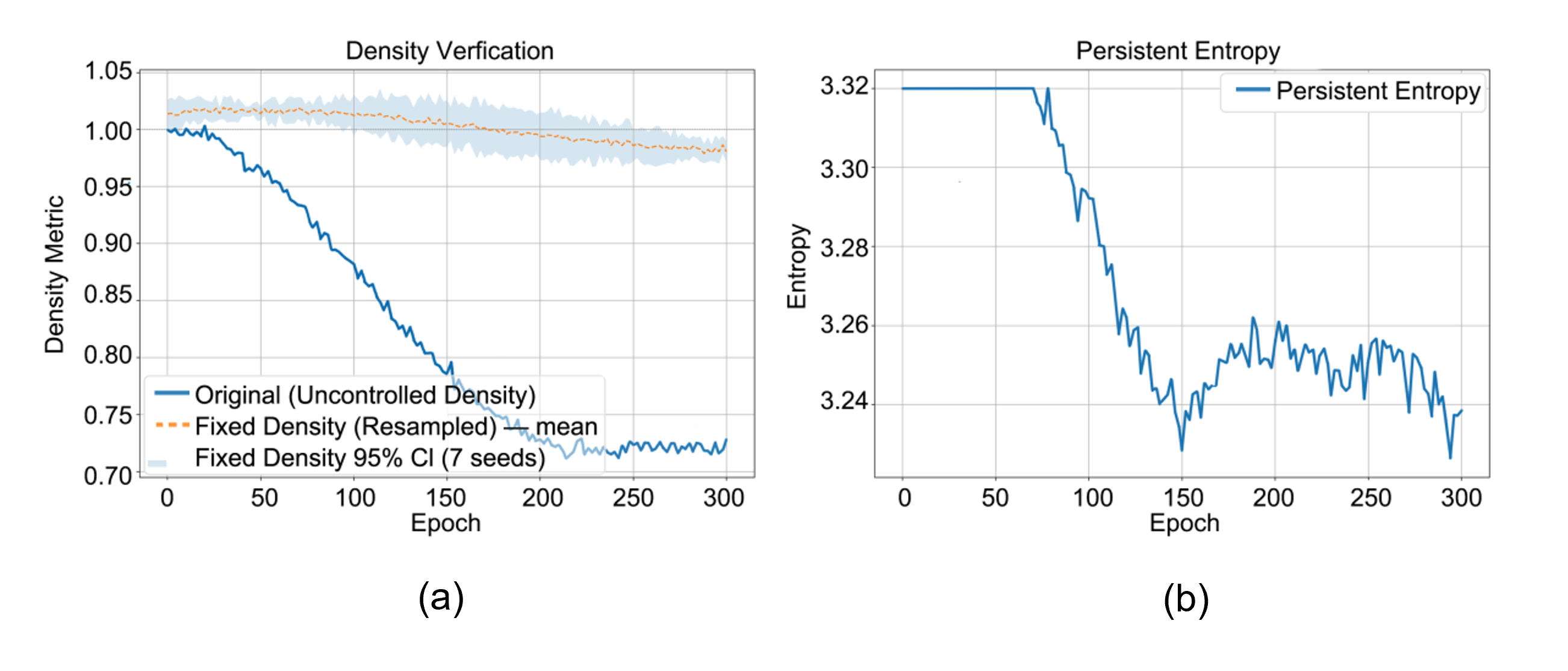}
  \caption{Changes in density and topological complexity of soft prompts during the training process.}
  \label{fig:2}
\end{figure}

Based on the aforementioned procedure, Figure \ref{fig:1} presents the results of multi-sample training on the GSM8K \cite{cobbe2021training} dataset, illustrating the evolution of soft prompts and the improvement of reasoning performance during LLM training. Subplots (a) the average persistence lifespan curve and (c) the topological feature count curve indicate that the number of $H_0$ features remains stable under both single-sample and multi-sample training. Meanwhile, the number of $H_1$ features decreases significantly during training, with faster and more stable reduction observed in multi-sample training. This trend is corroborated by (b) persistence diagrams and (d) persistence barcodes, where many $H_1$ features gradually vanish while $H_0$ features persist longer and become more prominent.

Figure \ref{fig:2} examines changes in density and topological complexity. In the left panel (a), after seven iterations, the range and mean values indicate that $H_1$ density decreases markedly during training, whereas the overall point cloud density exhibits only minimal reduction. This confirms that the decrease in $H_1$ is driven by redundancy elimination rather than an increase in global density. The right panel (b) shows persistence entropy rapidly declining from its initial high value before stabilizing. The moderate reduction reflects that although $H_1$ features decrease, a large number of $H_0$ features persist throughout training.


\begin{table}[h]
\centering
\small
\caption{Correlation and significance of topological features and accuracy}
\resizebox{\columnwidth}{!}{%
\begin{tabular}{lcccc}
\toprule
\multirow{2}{*}{\textbf{Metric}} & \multicolumn{2}{c}{\textbf{Spearman Correlation}} & \multicolumn{2}{c}{\textbf{Mann-Whitney U Test}} \\
\cmidrule(lr){2-3} \cmidrule(lr){4-5}
& $\rho$ & p-value & $U$ & p-value \\
\midrule
$|H_0|$ & N/A & N/A & $2664.5$ & $1.000$ \\
$|H_1|$ & $-0.324$ & $6.60 \times 10^{-5}$ & $3600.5$ & $9.7 \times 10^{-5}$ \\
Avg. $l_i$ ($H_0$) & $0.866$ & $3.40 \times 10^{-45}$ & $0.0$ & $1.9 \times 10^{-25}$ \\
Avg. $l_i$ ($H_1$) & $0.018$ & $0.826$ & $2608.0$ & $0.827$ \\
PE & $-0.809$ & $5.10 \times 10^{-35}$ & $5153.0$ & $2.1 \times 10^{-22}$ \\
\bottomrule
\end{tabular}%
}
\label{tab:topo_metrics}
\end{table}

To relate quantitative metrics to soft prompt tuning, we further analyze their correlations and significance. Table \ref{tab:topo_metrics} reports Spearman rank correlations \cite{sedgwick2014spearman} and Mann-Whitney U tests \cite{macfarland2016mann}, showing that applying trained soft prompts to LLMs yields a negative correlation between $H_1$ count and reasoning accuracy, indicating that models with higher performance contain fewer redundant loops. The average lifespan of $H_0$ features exhibits a strong positive correlation with accuracy, suggesting that more stable connected structures correspond to higher performance. Persistence entropy shows a strong negative correlation with accuracy, indicating that simplified topological structures correspond to improved model performance. The $H_0$ count remains constant in the validation examples, which is attributed to the use of low-variance Gaussian initialization that yields soft prompt embeddings forming a single connected component. This observation holds across all datasets (see in the supplementary materials). The analysis identifies $H_0$ average lifespan and persistence entropy as key metrics for assessing structural optimization and performance improvement. Overall, persistent homology metrics capture the evolution of soft prompts in the parameter space. The training process of soft prompt tuning maintains basic connectivity, while loops and redundant structures gradually disappear, leading to a more compact and stable vector structure. This indicates that training proceeds along an effective direction for optimizing LLM representations.

\section{Topological Soft Prompt Loss}
\label{sec:4}

\begin{figure*}[h!]
  \centering
  \includegraphics[width=\linewidth]{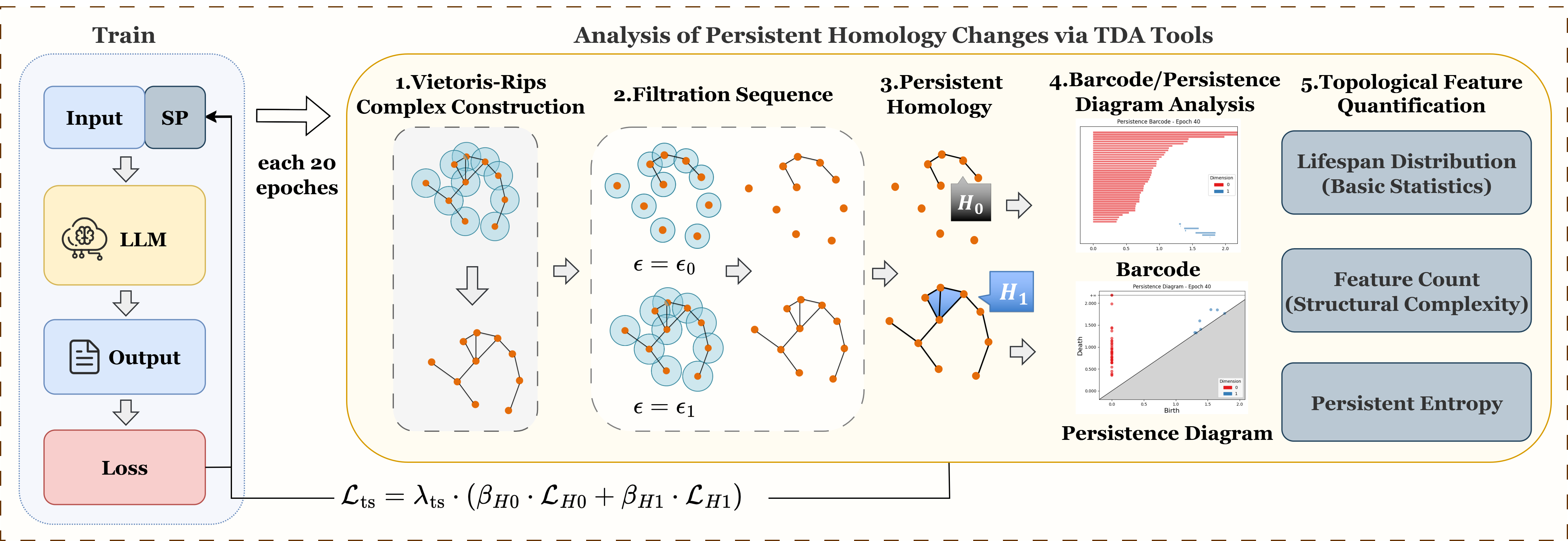}
  \caption{Overview of the structural analysis and optimization framework for soft prompts. Persistent homology is used to characterize the structural evolution of soft prompt representations during training, which motivates the design of the proposed TSLoss.}
  \label{fig:main}
\end{figure*}

Based on the analysis results, we design the \textbf{Topological Soft Prompt Loss (TSLoss)}. The loss regulates the topological structure of soft prompt vectors in the parameter space, ensuring that their representations remain stable and consistent. Specifically, based on significance and correlation analysis, TSLoss consists of two components: a loss based on $H_0$ features that maintains connectivity by controlling the soft nearest neighbor distance of each point to regulate local density, and a loss based on $H_1$ features that preserves the local topological structure through enforced attraction and repulsion between points.

Given a set of data points $X = \{x_1, x_2, ..., x_n\}$ in the embedding space, we first define the distance matrix $D_{ij} = \|x_i - x_j\|_2$ for $i,j = 1,\dots,P$. Since the traditional nearest neighbor distance is non-smooth, we designed a softmin function for gradient computation as 
\begin{equation}
\begin{split}
s_i = \text{softmin}(D_i, \tau) = -\tau \log \sum_{j=1}^{P} \exp\left(-\frac{D_{ij}}{\tau}\right),
\end{split}
\end{equation}
where $\tau$ is the temperature parameter of an LLM.

\textbf{The $H_0$ feature-based loss} ($\mathcal{L}_{H0}$) stabilizes the connected structure by controlling local density. The soft nearest neighbor distance $s_i$ provides a differentiable approximation of the death scale $d_i$ of point $x_i$ in the 0-dimensional homology group. Thus, we design:
\begin{equation}
\begin{split}
\label{eq:2}
\bar{s} = \frac{1}{P} \sum_{i=1}^{P} s_i, \quad \mathcal{L}_{H0} = \frac{1}{P} \sum_{i=1}^{P} (s_i - \bar{s})^2.
\end{split}
\end{equation}

Minimizing $H_0$ is equivalent to forcing all points to have similar local density, equivalent to minimizing the variance of the $H_0$ lifetime distribution, avoiding abnormally sparse or dense regions. This reflects the objective of maintaining consistent persistence of 0-dimensional features in the persistent homology diagram, as feature lifetime exhibits a significant correlation with model accuracy.

\textbf{The $H_1$ feature-based loss} ($\mathcal{L}_{H1}$) targets one-dimensional homology features by jointly modeling attraction and repulsion between point pairs to regulate ring-like structures. Specifically, for a given pairwise distance matrix $D_{ij}$, we define
\begin{equation}
\begin{split}
\label{eq:3}
\mathcal{L}_{H1}= \frac{1}{n^2} \sum_{i,j} \Big[\lambda_{\text{repel}} \max(0, \delta - D_{ij})^2 \\
+\lambda_{\text{attract}} \max(0, D_{ij} - \zeta)^2\Big].
\end{split}
\end{equation}

The first term in EQ. \eqref{eq:3} penalizes overly close point pairs below the local scale $\delta$, suppressing spurious edges and preventing premature filling of loops, while the second term penalizes excessively distant pairs beyond the global scale $\zeta$, preserving sufficient connectivity to avoid ring fragmentation. Together, these terms constrain the emergence and dissolution of $H_1$ features to appropriate distance ranges during Vietoris-Rips complex construction, thereby stabilizing meaningful ring structures while eliminating topological redundancy.

\textbf{Soft quantile threshold calculation} approximates the multi-scale behavior of TDA by estimating representative local and global distance thresholds via soft quantiles. Given the pairwise distance matrix $D_{ij}$, we define
\begin{equation}
\begin{split}
\delta = \sum_{i,j} w^{\text{low}}_{ij} \cdot D_{ij} ; \zeta = \sum_{i,j} w^{\text{high}}_{ij} \cdot D_{ij},
\end{split}
\end{equation}
with
\begin{equation}
\begin{split}
w^{\text{low}}_{ij} = \frac{e^{-\alpha D_{ij}}}{\sum_{k,l} e^{-\alpha D_{kl}}};w^{\text{high}}_{ij} = \frac{e^{\alpha D_{ij}}}{\sum_{k,l} e^{\alpha D_{kl}}}.
\end{split}
\end{equation}
where $\delta$ and $\zeta$ serve as soft estimates of the lower and upper quantiles of the distance distribution, corresponding to local and global scales, respectively. The parameter $\alpha$ controls the sharpness of the approximation, recovering hard quantile behavior as $\alpha \to \infty$.

Combining the above, the complete topological constancy loss function is defined as
\begin{equation}
\begin{split}
\mathcal{L}_{\text{ts}} = \lambda_{\text{ts}} \cdot (\beta_{H0} \cdot \mathcal{L}_{H0} + \beta_{H1} \cdot \mathcal{L}_{H1}),
\end{split}
\end{equation}
where $\lambda_{\text{ts}}$ represents the overall weight of TSLoss, and $\beta_{H0}$ and $\beta_{H1}$ control the relative contributions of $H_0$ and $H_1$ terms respectively. The normalization of distance and density terms in Eq. \eqref{eq:2} and \eqref{eq:3} maintains numerical consistency between the two sub-losses, rendering TSLoss parameter-insensitive.

Since this function does not independently quantify differences between predicted and true label distributions, TSLoss serves as an auxiliary regularizer combined with other loss functions such as cross-entropy loss \cite{mao2023cross}. Under the persistent homology framework, this loss function incorporates conclusions from Section \ref{sec:2} by learning associative features of 0-dimensional and 1-dimensional homology groups during soft prompt training. This approach results in trained soft prompts that are topologically more stable, directing intrinsic geometric structure evolution and thereby improving representation learning quality and convergence speed.

\section{Experiments}
\label{sec:5}
In this section, we present all detailed settings, models, and datasets used in our validation experiments for the structural analysis, demonstrating the effectiveness of TSLoss. To maintain logical narrative flow and showcase the design rationale for TSLoss, we choose to present the topological analysis data in Section \ref{sec:2}, while providing supplementary details about experimental configurations here.

\subsection{Datasets, Models, and Baselines}
\label{subsec:1}


To ensure the generality of the evaluation, both the structural analysis and the effectiveness of TSLoss optimization are evaluated on three benchmark datasets: GSM8K \cite{cobbe2021training}, which focuses on elementary mathematics word problems, MMLU-CF \cite{zhao2024mmlu}, which targets common-sense and factual reasoning, and LongBench \cite{bai2023longbench}, which evaluates long-context reasoning. These datasets cover a broad range of task types, from basic arithmetic to multi-hop reasoning and problem solving.

\begin{table*}[h!]
\centering
\caption{Accuracy (\%) of different soft prompt tuning methods across models and datasets.}
\label{tab:soft_prompt_comparison}
\resizebox{\textwidth}{!}{
\begin{tabular}{llccccccc}
\toprule
\textbf{Model} & \textbf{Dataset} &
\textbf{Soft Prompt} &
\textbf{+ L2-Reg} &
\textbf{+ PDLoss} &
\textbf{ACT} &
\textbf{Prefix-Tuning} &
\textbf{DPC} &
\textbf{Ours} \\
\midrule

\multirow{3}{*}{Gemma-2B-IT}
& GSM8K     & 14.2 & 14.8 & 16.5 & 16.0 & 17.8 & 19.8 & \textbf{20.5} \\
& MMLU      & 26.5 & 27.0 & 28.8 & 28.2 & 29.5 & 31.8 & \textbf{32.4} \\
& LongBench & 15.1 & 15.5 & 17.2 & 16.8 & 18.0 & 20.0 & \textbf{20.8} \\
\midrule

\multirow{3}{*}{Open-LLaMA-7B}
& GSM8K     & 32.5 & 33.1 & 35.8 & 34.2 & 36.5 & 37.9 & \textbf{38.2} \\
& MMLU      & 40.2 & 40.8 & 42.5 & 41.5 & 43.1 & \textbf{45.2} & 44.8 \\
& LongBench & 21.5 & 21.8 & 23.2 & 22.5 & 23.8 & 24.8 & \textbf{25.1} \\
\midrule

\multirow{3}{*}{DeepSeek-7B-Chat}
& GSM8K     & 48.2 & 49.0 & 51.5 & 50.8 & 52.1 & 54.0 & \textbf{54.6} \\
& MMLU      & 52.5 & 52.8 & 54.2 & 53.5 & 55.0 & 56.5 & \textbf{56.8} \\
& LongBench & 28.1 & 28.4 & 29.8 & 29.0 & 30.2 & 31.4 & \textbf{31.8} \\
\midrule

\multirow{3}{*}{LLaMA2-13B}
& GSM8K     & 58.5 & 59.2 & 61.5 & 60.8 & 62.0 & 64.0 & \textbf{64.8} \\
& MMLU      & 58.0 & 58.5 & 60.0 & 59.5 & 61.0 & 62.8 & \textbf{63.0} \\
& LongBench & 32.5 & 33.0 & 34.5 & 34.0 & 35.2 & 36.5 & \textbf{37.2} \\
\midrule

\multirow{3}{*}{Qwen1.5-14B}
& GSM8K     & 72.0 & 72.5 & 74.0 & 73.5 & 74.8 & 76.4 & \textbf{77.2} \\
& MMLU      & 68.0 & 68.5 & 70.0 & 69.5 & 71.0 & 72.8 & \textbf{73.0} \\
& LongBench & 40.0 & 40.5 & 42.0 & 41.5 & 43.0 & 44.5 & \textbf{45.4} \\
\bottomrule
\end{tabular}
}
\end{table*}

Five LLMs are selected for evaluating TSLoss: DeepSeek-7B-Chat \cite{bi2024deepseek}, Open-LLaMA-7B \cite{geng2023openllama}, Gemma-2B-IT, LLaMA2-13B \cite{touvron2023llama}, and Qwen1.5-14B \cite{bai2023qwen}. Structural analysis is conducted only on Gemma-2B-IT, as prompt tuning on larger models is generally less effective and more resource-intensive, with performance improvements dominated by parameter scale. Accordingly, models at the 2B, 7B, and 14B scales are considered, where smaller models exhibit less stable prompt responses and therefore provide a more informative setting for structural analysis.

To assess the effectiveness of the proposed approach, the baseline soft prompt tuning method \cite{lester2021power} is compared with variants incorporating L2-regularization (L2-Reg) and pairwise distance loss (PDLoss). In addition, comparisons are conducted with Prefix-Tuning \cite{li2021prefix}, P-Tuning v2 \cite{liu2022p}, ACT \cite{yu2024unveiling}, and DPC \cite{fan2025improving}. 

\subsection{Implementation Details}
\label{subsec:2}
To systematically analyze structural variations of soft prompts across scenarios, we consider single-sample and multi-sample training settings. These configurations are used for both the topological phenomenon analysis and the evaluation of TSLoss optimization capability. To avoid interference from data distribution differences, both training settings draw samples from the same dataset, ensuring controllability and consistency in structural feature comparison. In the multi-sample setting, each task is trained with 10 samples and evaluated on 100 test instances per dataset.

All training processes adopt identical initialization methods, optimizers, and hyperparameters across different task settings. Specifically, soft prompt vectors are initialized from a Gaussian distribution ($\mathcal{N}(0, 0.02^2 \mathbf{I})$), and training employs the AdamW optimizer with a learning rate of $5 \times 10^{-5}$, $\lambda_{\text{ts}}=1$, weight decay $0.01$, $\epsilon = 10^{-8}$, batch size 8, and 300 training epochs. For TSLoss, we set $\beta_{H0}=\beta_{H1}=1$ to assign equal importance to connectivity preservation and local loop structure regularization. A linear learning rate scheduler with a 10\% warm-up phase is used. During training, topological data analysis is performed every 20 epochs, while inference accuracy is recorded to characterize the evolutionary process of soft prompts. Representative samples illustrating key topological feature changes throughout training are presented, with comprehensive comparisons provided in the supplementary materials. All experiments are conducted on a computing environment equipped with an Intel Xeon Platinum 8173M CPU (2.00GHz, 28 cores, 112 threads) and eight NVIDIA GeForce RTX 3090 GPUs (24GB each). In the evaluation of TSLoss optimization capability, the total loss function is defined as
$L_{\text{total}} = L_{\text{ce}} + \lambda_{\text{ts}} \cdot L_{\text{ts}}$, where $L_{\text{ce}}$ denotes the cross-entropy loss.

\subsection{Main Result}
Table~\ref{tab:soft_prompt_comparison} summarizes the accuracy of different soft prompt tuning methods in models and datasets. Overall, the TSLoss-based approach achieves the best or near-best performance in most cases. Compared to L2-regularization and pairwise distance loss, the proposed method yields superior performance. This improvement highlights that topological regularization enforces structural consistency in soft prompt representations.

For models with lower parameter counts, the performance gains are most substantial. On GSM8K, Gemma-2B-IT improves from 19.8 to 20.5, and similar trends are observed on MMLU and LongBench. These results indicate that topological regularization is particularly effective in lower-capacity settings, where soft prompt representations are more sensitive to optimization instability. Performance improvements are also observed for larger models. For instance, Qwen1.5-14B reaches 77.2 on GSM8K, outperforming all baseline methods. Although the absolute gains decrease with increasing model scale, the results demonstrate that the proposed method remains effective across models of varying parameter scales. These findings collectively suggest that incorporating topological constraints into soft prompt tuning facilitates more stable optimization and enhances generalization across different models and tasks, while maintaining consistent effectiveness across model scales.
\begin{table*}[h!]
\centering
\small
\setlength{\tabcolsep}{6pt}  
\caption{Number of iterations for soft prompts to achieve complete task accuracy.}
\begin{tabular}{llcccccc}
\hline
\multirow{2}{*}{Model} & \multirow{2}{*}{Method} & \multicolumn{3}{c|}{Single-sample} & \multicolumn{3}{c}{Multi-sample} \\
\cline{3-8}
& & GSM8K & MMLU-CF & LongBench & GSM8K & MMLU-CF & LongBench \\
\hline
\multirow{4}{*}{\shortstack[l]{Gemma\\2B-IT}}
& Standard      & 154 & 8 & 12 & 118 & 16 & 104 \\
& Prefix-Tuning & 128 & 7 & 10 & 96  & 15 & 88  \\
& P-Tuning v2   & 96  & 6 & 9  & 74  & 14 & 65  \\
& \textbf{Ours} & \textbf{88} & \textbf{6} & \textbf{8} & \textbf{62} & \textbf{14} & \textbf{58} \\
\hline
\multirow{4}{*}{\shortstack[l]{Open-LLaMA\\7B}}
& Standard      & 12  & 4 & 6  & 10  & 8  & 38  \\
& Prefix-Tuning & 11  & 3 & 5  & 9   & 6  & 36  \\
& P-Tuning v2   & 10  & 2 & 4  & 8   & 5  & 35  \\
& \textbf{Ours} & \textbf{10} & \textbf{2} & \textbf{4} & \textbf{8} & \textbf{4} & \textbf{34} \\
\hline
\multirow{4}{*}{\shortstack[l]{DeepSeek\\7B-Chat}}
& Standard      & 8   & 2 & 4  & 8   & 4  & 32  \\
& Prefix-Tuning & 7   & 2 & 3  & 7   & 3  & 30  \\
& P-Tuning v2   & 6   & 2 & 2  & 5   & 2  & 29  \\
& \textbf{Ours} & \textbf{6} & \textbf{2} & \textbf{2} & \textbf{4} & \textbf{2} & \textbf{28} \\
\hline
\multirow{4}{*}{\shortstack[l]{LLaMA2\\13B}}
& Standard      & 9   & 3 & 5  & 9   & 6  & 30  \\
& Prefix-Tuning & 8   & 3 & 4  & 8   & 5  & 28  \\
& P-Tuning v2   & 7   & 2 & 3  & 6   & 4  & 26  \\
& \textbf{Ours} & \textbf{7} & \textbf{2} & \textbf{3} & \textbf{5} & \textbf{4} & \textbf{25} \\
\hline
\multirow{4}{*}{\shortstack[l]{Qwen1.5\\14B}}
& Standard      & 6   & 2 & 3  & 7   & 3  & 24  \\
& Prefix-Tuning & 5   & 2 & 3  & 6   & 3  & 22  \\
& P-Tuning v2   & 4   & 1 & 2  & 5   & 2  & 20  \\
& \textbf{Ours} & \textbf{4} & \textbf{1} & \textbf{2} & \textbf{4} & \textbf{2} & \textbf{19} \\
\hline
\end{tabular}
\label{tab:training_iterations}
\end{table*}

\subsection{Convergence Analysis}
In this experiment, we measure the number of training iterations required for soft prompts to achieve 100\% accuracy on task-specific problem instances when fine-tuned for each LLM. Table~\ref{tab:training_iterations} reports the iterations needed under single-sample and multi-sample training configurations. Across all LLMs and datasets, the proposed method achieves faster convergence compared with Standard Tuning, Prefix-Tuning, and P-Tuning v2. The reduction in the number of iterations is more substantial for lower-capacity models. For example, Gemma-2B-IT on GSM8K decreases from 118 to 62 iterations under the multi-sample setting. Higher-capacity models also exhibit accelerated convergence, with Qwen1.5-14B on GSM8K reducing iterations from 7 to 4. These results indicate that integrating topological regularization enhances training efficiency across models of varying scales.

\begin{figure}[h!]
  \centering
  \includegraphics[width=\linewidth]{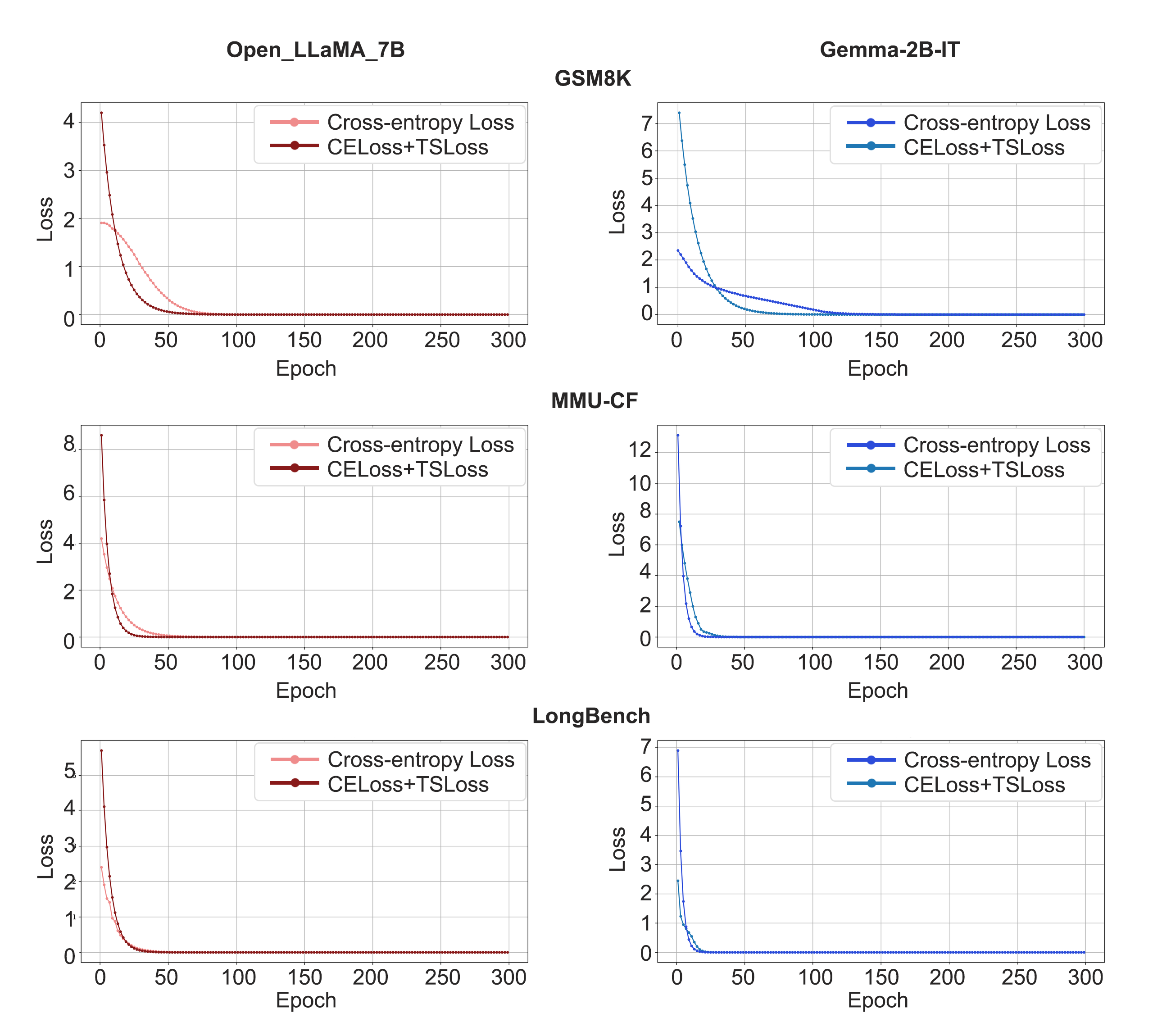}
  \caption{Comparison of convergence behavior of overall loss functions with and without TSLoss across different datasets and models.}
  \label{fig:3}
\end{figure}

The loss curves in Figure \ref{fig:3} illustrate the training dynamics of soft prompts with and without TSLoss across three datasets and two LLMs. In all cases, combining cross-entropy loss with TSLoss accelerates convergence, reducing the initial loss more rapidly and stabilizing earlier. The effect is particularly pronounced for the smaller Gemma-2B-IT model, where TSLoss significantly lowers the early training loss compared to cross-entropy alone, indicating improved optimization stability.


\subsection{Parameter Sensitivity Analysis}
\begin{figure}[h!]
  \centering
  \includegraphics[width=\linewidth]{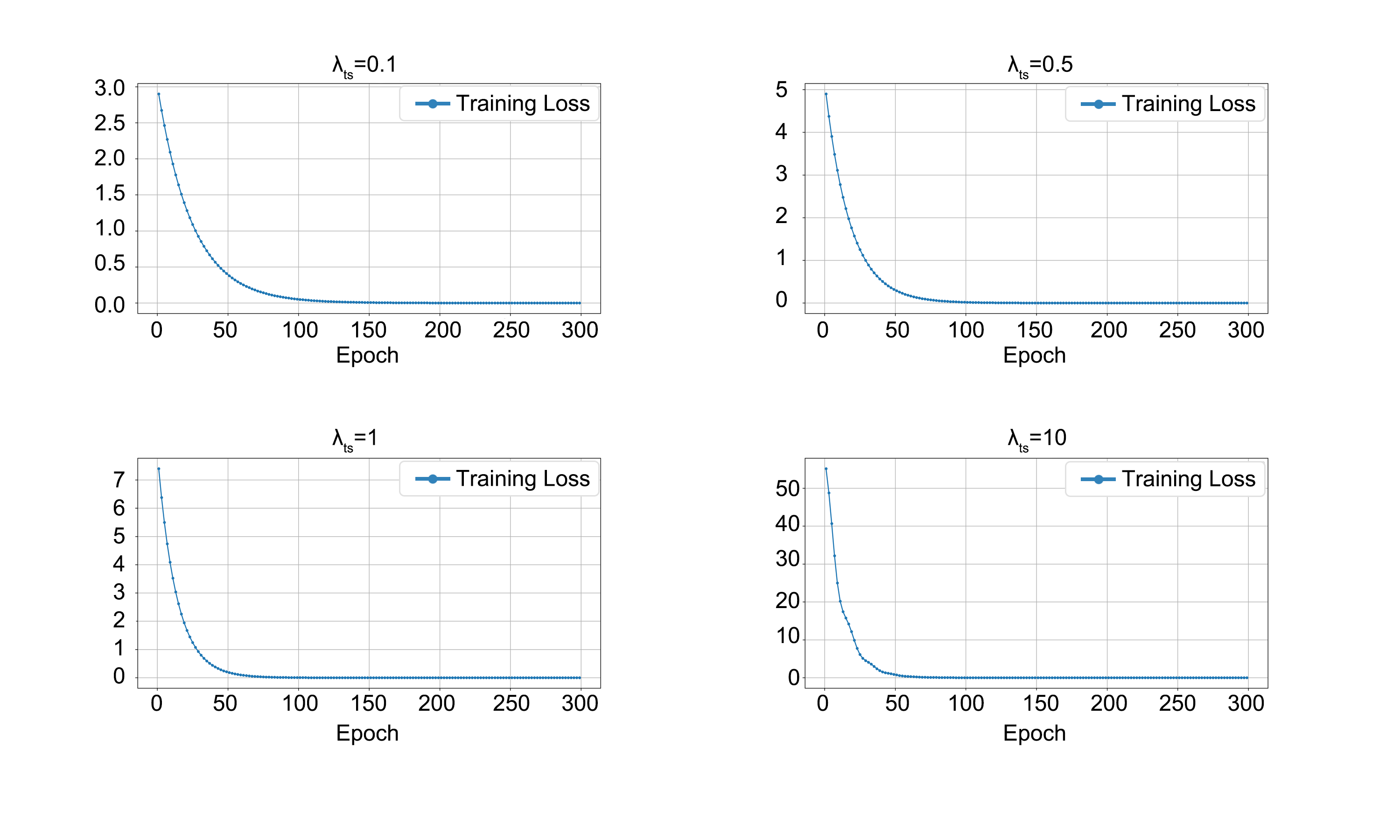}
  \caption{Convergence of overall loss under different TSLoss weight values.}
  \label{fig:4}
\end{figure}

Figure~\ref{fig:4} illustrates the convergence behavior of the total loss $L_{total}$ under different $\lambda_{ts}$ settings. All configurations reach $L_{total} < 0.1$ within approximately 100 epochs on average, demonstrating robust convergence across a broad range of $\lambda_{ts}$ values. Moderate values of $\lambda_{ts}$ (0.1–1) result in faster convergence, whereas excessively large values (e.g., $\lambda_{ts}=10$) induce oscillations and prevent convergence, likely due to the overemphasis on structural constraints that hinder correction of initial embedding inconsistencies. Table~\ref{tab:lambda_alpha_iterations} reports the number of iterations required for convergence under different $\lambda_{ts}$ and temperature settings. Similarly, the temperature parameter exhibits sensitivity, with intermediate values accelerating convergence and larger values slowing it down.

\begin{table}[h]
\centering
\caption{Impact of weight ($\lambda$) and temperature ($\alpha$) on training iterations.}
\begin{tabular}{cc|cc}
\toprule
\multicolumn{2}{c}{Weight ($\lambda$)} & \multicolumn{2}{c}{Temperature ($\alpha$)} \\
\cmidrule(lr){1-2}\cmidrule(lr){3-4}
Value & Iter. & Value & Iter. \\
\midrule
0   & 154 & 5  & 112 \\
0.1 & 130 & 10 & 88  \\
0.5 & 112 & 20 & 135 \\
1   & 88  & -- & --  \\
10  & --  & -- & --  \\
\bottomrule
\end{tabular}
\label{tab:lambda_alpha_iterations}
\end{table}

\subsection{Generalization Analysis}

In this section, we analyze the generalization of TSLoss by comparing different ablation settings against the cross-entropy baseline on GSM8K under the single-sample setting in Table \ref{tab:convergence_components}. As designed in Section \ref{sec:4}, incorporating the regularization components $H_0$ and $H_1$ progressively reduces the number of iterations required for convergence, enabling the model to better capture structural patterns in the data. The full version of TSLoss achieves the fastest convergence and the largest relative improvement, highlighting its distinctive effect compared to CELoss.

\begin{table}[h!]
\centering
\footnotesize
\setlength{\tabcolsep}{3pt}
\caption{Convergence iterations and relative improvement of different loss components on GSM8K under the single-sample setting.}
\begin{tabular}{l l c c}
\toprule
Model & Configuration & Iterations to Converge & Improvement \\
\midrule
\multirow{4}{*}{\shortstack[l]{Gemma-2B-IT}}
& CE Baseline        & 154 & --      \\
& CE + $H_0$ only    & 132 & 14.30\% \\
& CE + $H_1$ only    & 115 & 25.30\% \\
& CE + TSLoss  & 88  & 42.90\% \\
\midrule
\multirow{4}{*}{\shortstack[l]{Open-LLaMA-7B}}
& CE Baseline        & 12  & --      \\
& CE + $H_0$ only    & 11  & 8.30\%  \\
& CE + $H_1$ only    & 11  & 8.30\%  \\
& CE + TSLoss & 10  & 16.70\% \\
\bottomrule
\end{tabular}
\label{tab:convergence_components}
\end{table}

\begin{figure}[h!]
  \centering
  \includegraphics[width=\linewidth]{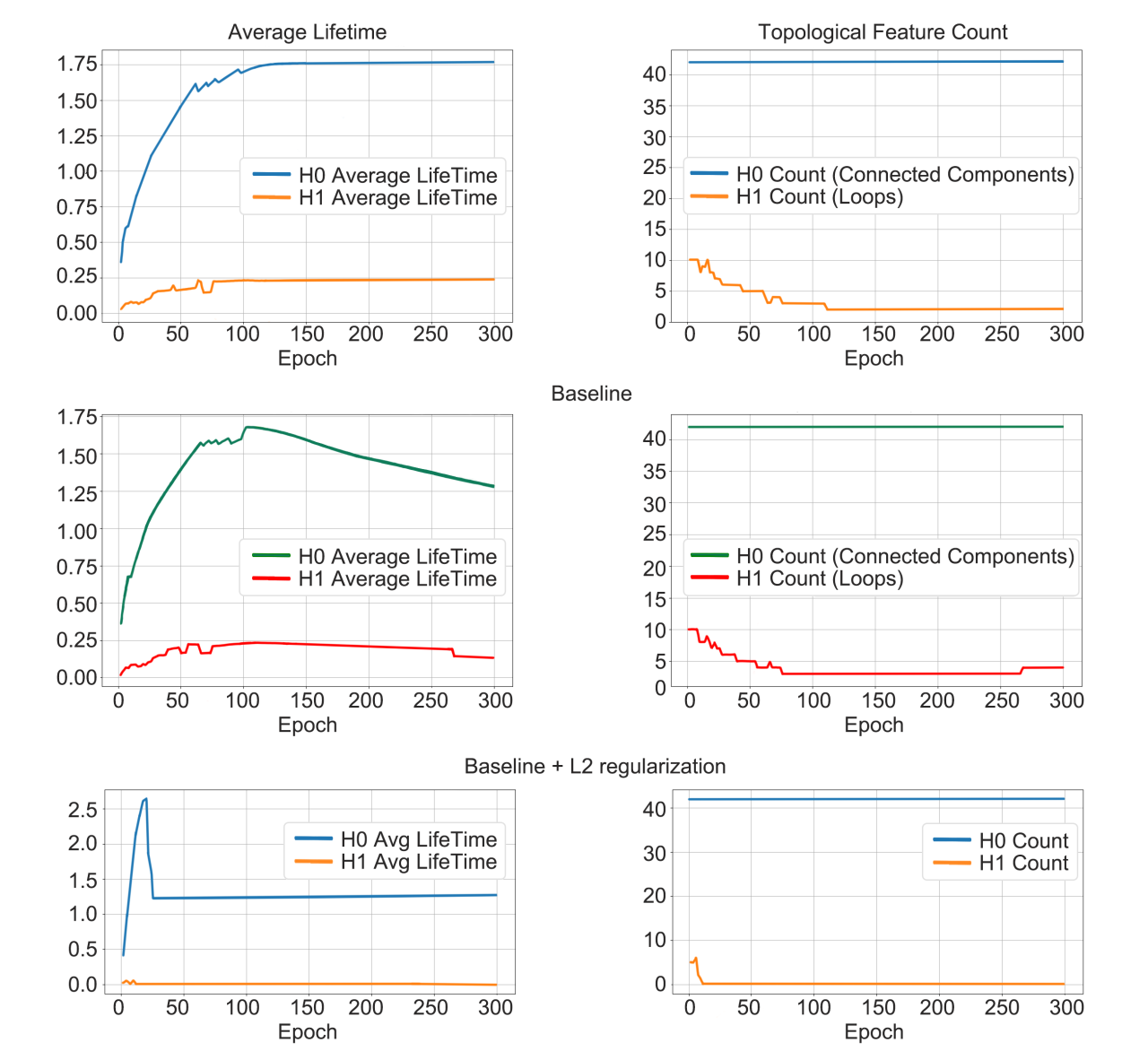}
  \caption{Evolution of topological features of soft prompts under different training strategies.}
  \label{fig:5}
\end{figure}

\subsection{Visualization Analysis}
\label{sec5.7}
In this section, we examine the evolutionary structural characteristics of soft prompts under different training strategies. Specifically, we present a set of visual analyses comparing the baseline soft prompt tuning approach, an L2-regularized variant, and TSLoss. These analyses include dimensionality reduction and topological data analysis, aiming to provide empirical support for the theoretical formulation and to evaluate the behavior of TSLoss.

\begin{figure}[h!]
  \centering
  \includegraphics[width=\linewidth]{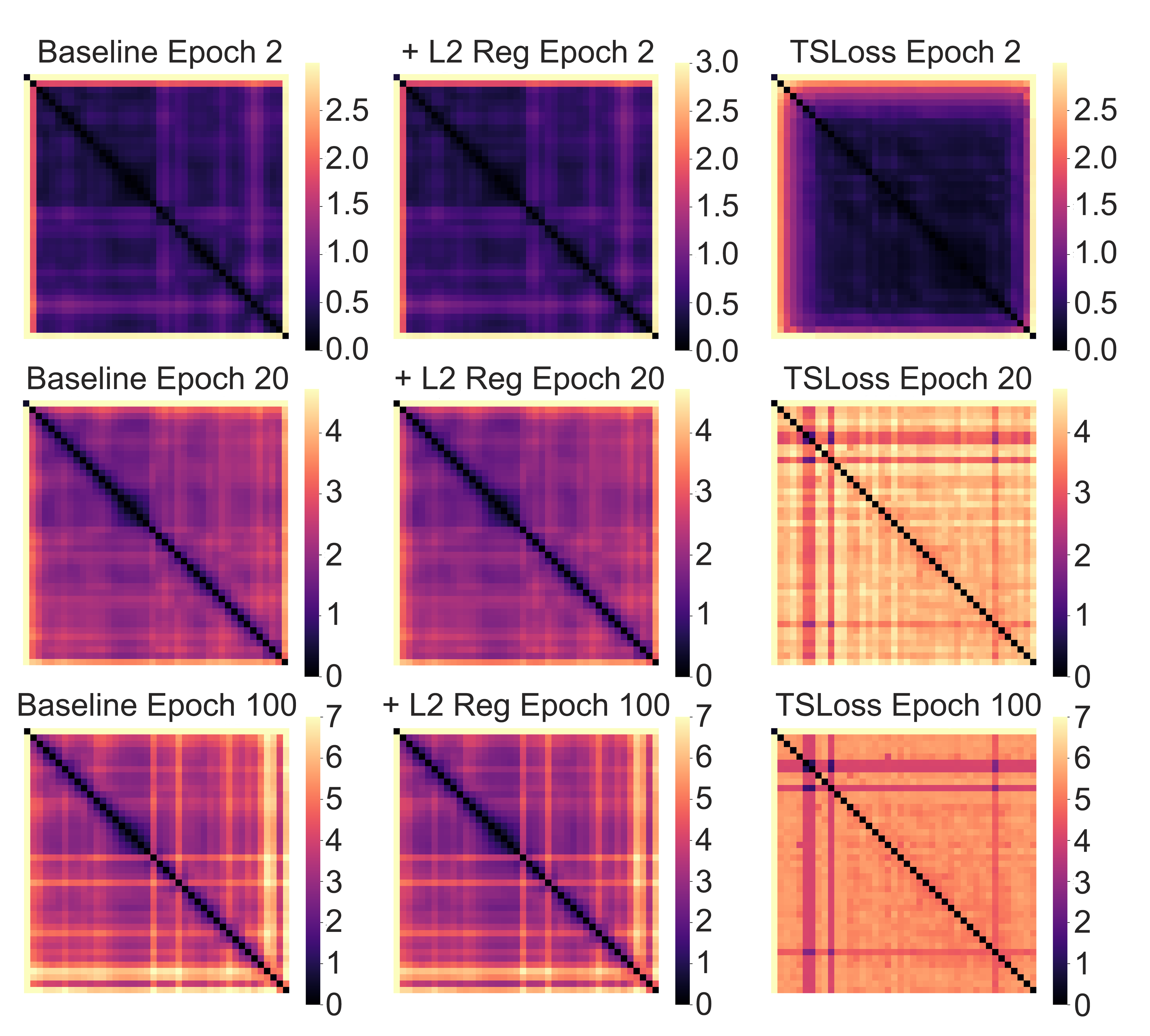}
  \caption{Heatmaps of the pairwise Euclidean distance matrix among soft prompt vectors.}
  \label{fig:8}
\end{figure}

Figure \ref{fig:5} illustrates the evolution of the number of persistent homology features and their average lifetimes across training. The baseline method exhibits considerable fluctuations in topological characteristics during later training stages. Incorporating L2-regularization results in a moderate improvement in structural stability. In comparison, the TSLoss-based strategy is associated with earlier onset and more sustained maintenance of topological features, indicating a more structured evolution of soft prompts.
\begin{figure*}[h!]
  \centering
  \includegraphics[width=0.77\linewidth]{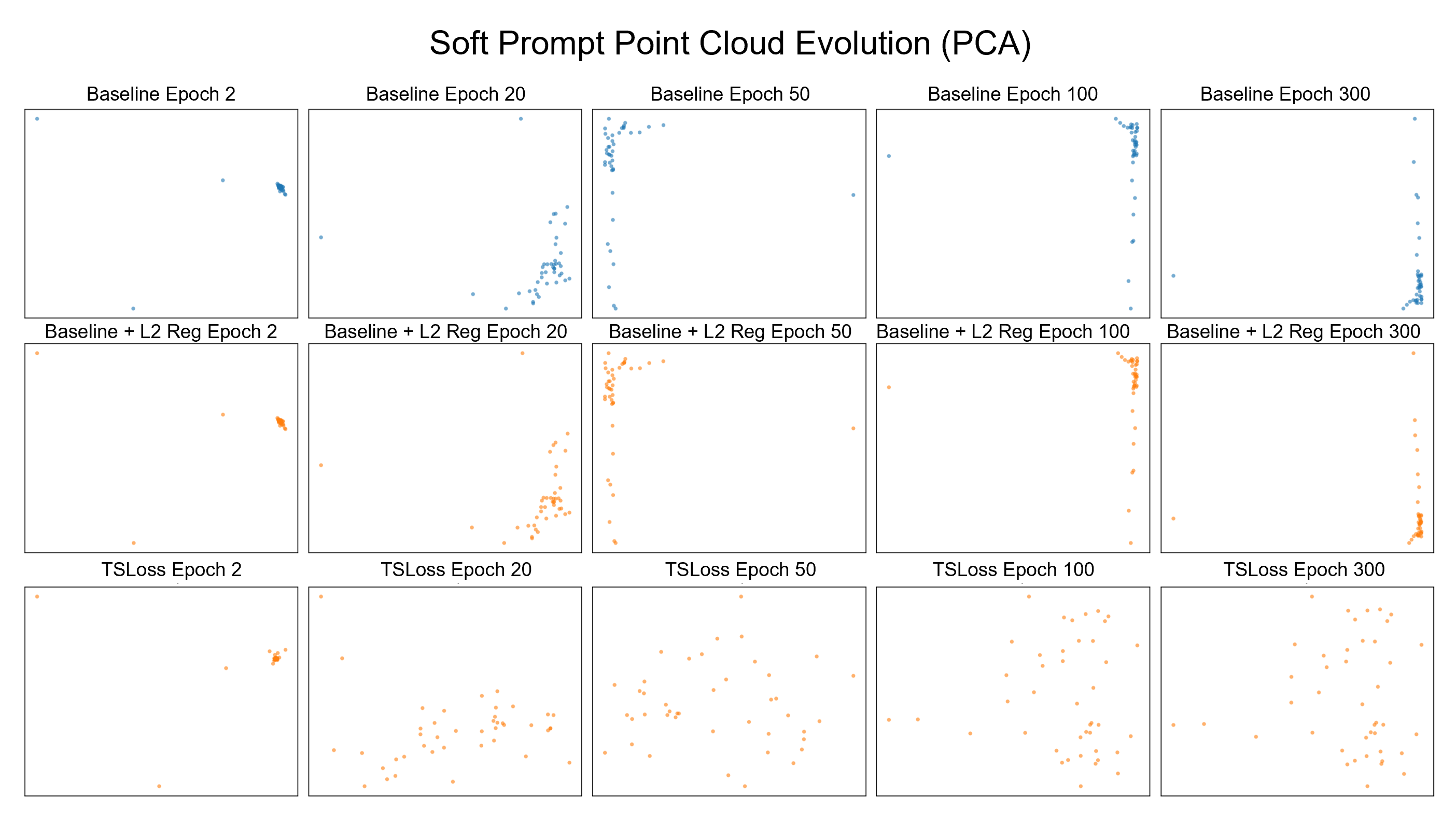}
  \caption{Evolution of soft prompt point clouds in PCA space.}
  \label{fig:6}
\end{figure*}
\begin{figure*}[h!]
  \centering
  \includegraphics[width=0.77\linewidth]{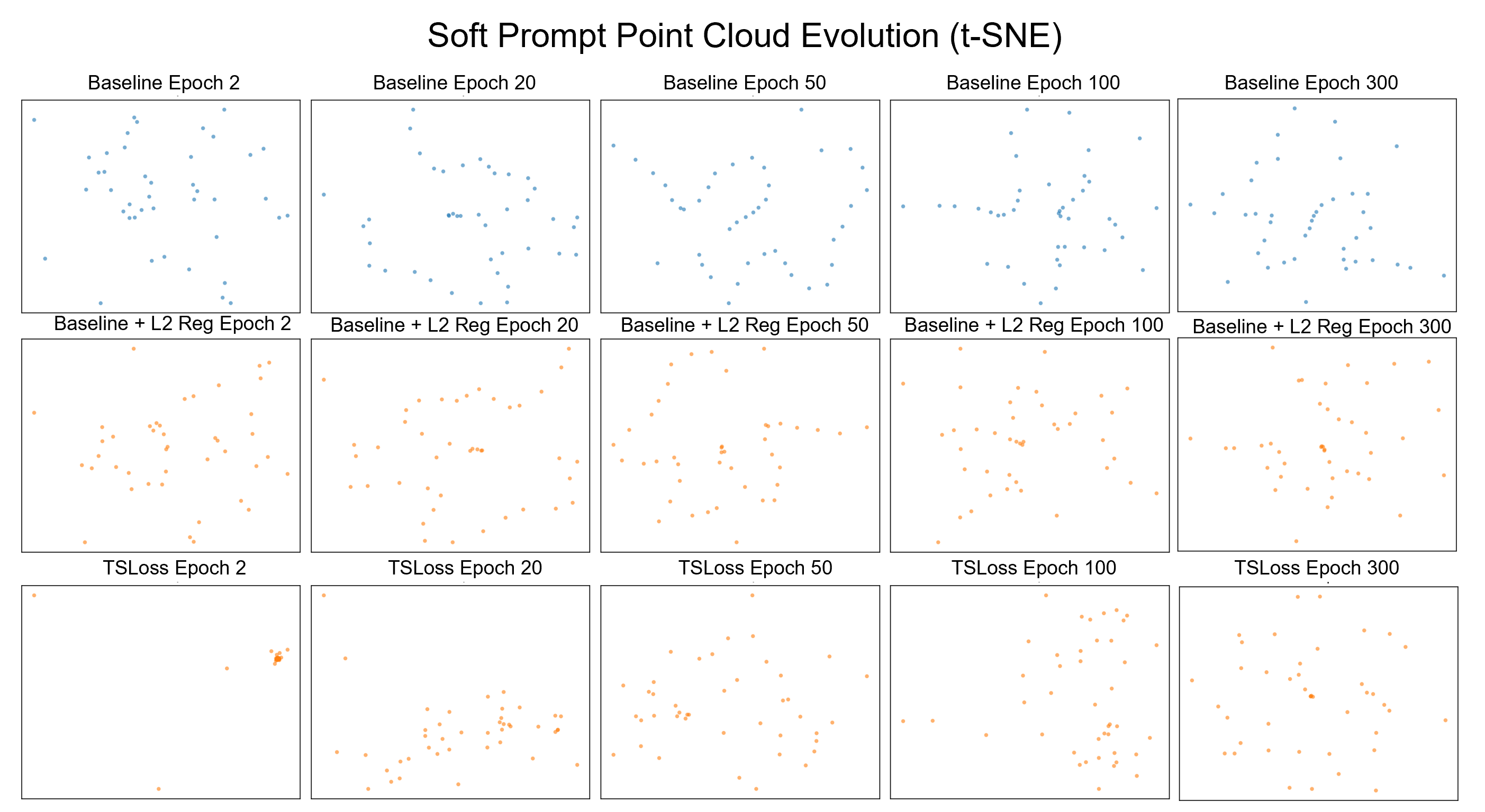}
  \caption{Evolution of soft prompt point clouds in t-SNE space.}
  \label{fig:7}
\end{figure*}

We compute Euclidean distance matrices of prompt vectors at multiple training stages and visualize them as heatmaps in Figure \ref{fig:8}. These results depict the evolution of prompt representations from random initialization to convergence. The color intensity represents pairwise distances between prompt vectors. Under the TSLoss strategy, pronounced block-like patterns emerge in the mid-to-late stages of training, indicating that prompts with similar semantics tend to cluster in representation space, while dissimilar prompts remain relatively separated. By contrast, the baseline method produces more diffuse distance patterns with limited structural organization. The L2-regularized approach primarily constrains vector magnitudes, but limits influence on semantic clustering. 

In addition, Figures \ref{fig:6} and \ref{fig:7} present visualizations of soft prompt vectors at different training stages using principal component analysis (PCA) \cite{abdi2010principal} and t-SNE \cite{maaten2008visualizing}, respectively. The PCA projections indicate that, as training progresses, soft prompts gradually shift from an initially dispersed distribution to more organized configurations in latent space. The t-SNE visualizations further reveal local neighborhood relationships among prompts and indicate increased separation between different prompt groups. Comparisons across training strategies suggest that TSLoss improves the compactness of prompt distributions and enhances inter-group separation.

\subsection{Variations of $H_0$ and $H_1$ counts}
\begin{figure}[h!]
    \centering
    \includegraphics[width=\linewidth]{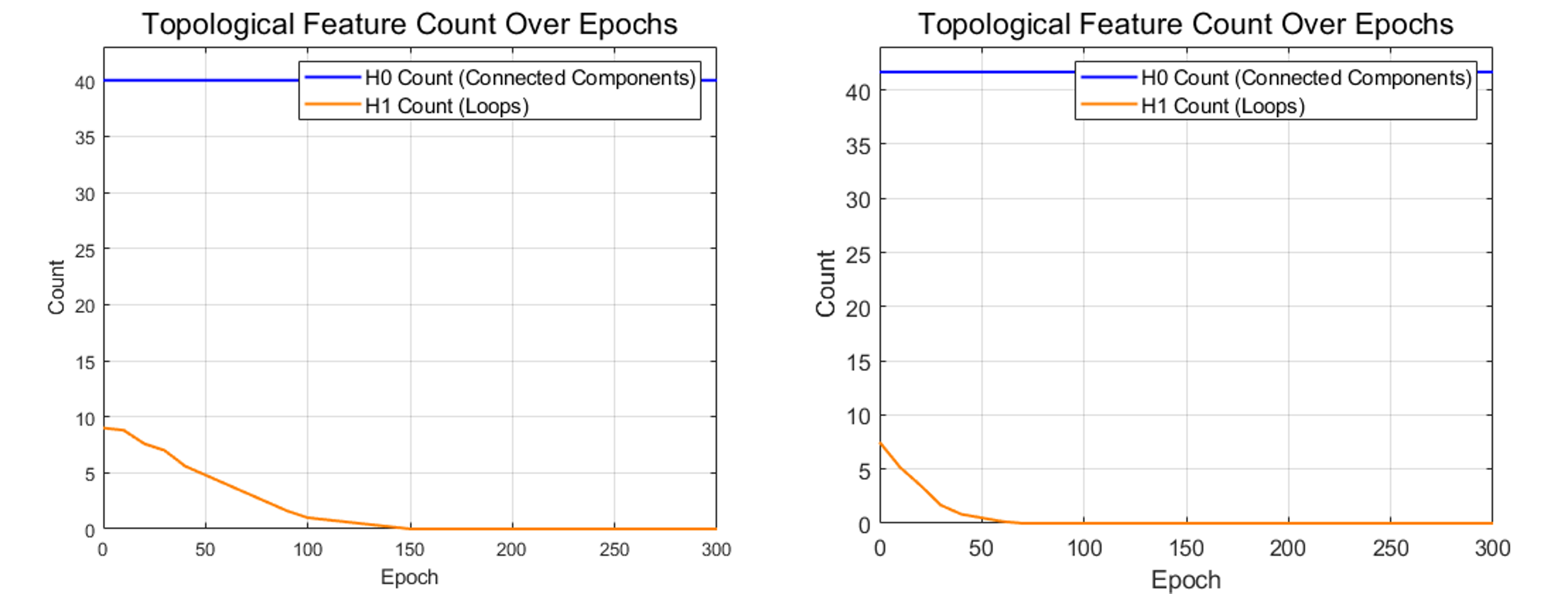}
    \caption{Averaged results of persistent homology in single-sample and multi-sample training on all datasets.}
    \label{fig:stat}
\end{figure}

Following the same settings in Section \ref{subsec:2}, we conducted a systematic statistical analysis across the diverse datasets to validate the reliability and generalizability of our conclusions. Specifically, we compare and average the quantities of $H_0$ and $H_1$ features and the changes in persistent entropy during both single-sample and multi-sample training processes. Figure \ref{fig:stat} shows the averaged results across all six datasets, where the left panel displays results from single-sample visualizations, and the right panel shows multi-sample results. The results in Figure \ref{fig:stat} reveal that, despite differences in data distribution and task context, the model exhibits consistent topological evolution trends during the training of soft prompts, further reinforcing our core hypothesis regarding structural evolution patterns.

\subsection{Lifetimes of Topological Features}
\begin{figure}[h!]
    \centering
    \includegraphics[width=0.9\linewidth]{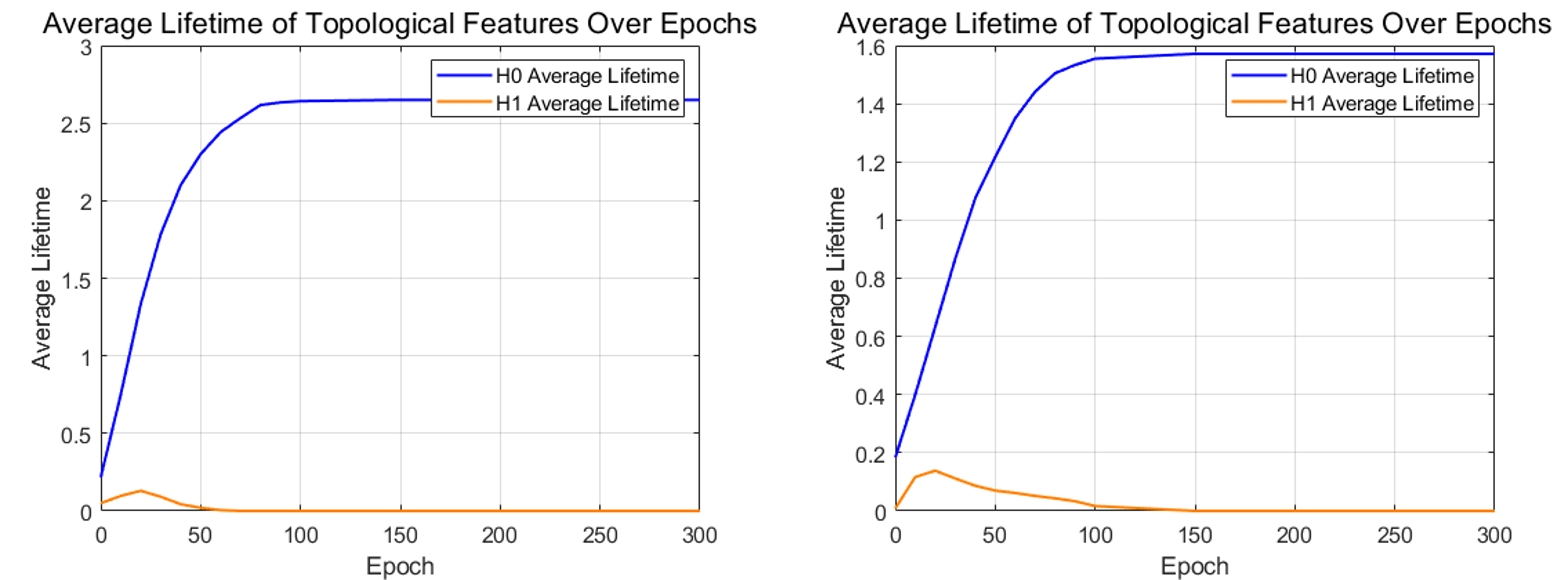}
    \caption{Averaged results of persistent homology barcodes and lifetime distributions across all datasets. Left panel: single-sample training; right panel: multi-sample training.}
    \label{fig:lifetime}
\end{figure}

Similarly, as shown in Fig. \ref{fig:lifetime}, the left and right panels depict the evolution of the average lifetimes of topological features across training iterations under single-sample and multi-sample training settings, respectively. The results indicate that the average lifetime of $H_0$ exhibits a monotonic increase and rapidly converges to a stable value in both training modes, highlighting that this structurally simple form is established early in training. In contrast, the average lifetime of $H_1$ shows a transient peak at the initial stage and quickly decays to baseline levels, suggesting that higher-order topological structures are gradually eliminated during model optimization. These findings provide a topological perspective on the dynamic process of soft prompt tuning, emphasizing the dominance and importance of simple and stable structures in the optimization.

\section{Conclusion}
In this work, we investigate the structural evolution of soft prompt representations during training process, providing a topological perspective on prompt interpretability. Using persistent homology from topological data analysis (TDA), we quantify the emergence of structurally stable and compact representations linked to effective tuning. Building on these findings, we propose the Topological Soft Prompt Loss (TSLoss), which regularizes prompt embeddings toward simplified yet well-connected structures by suppressing redundant patterns while preserving essential connectivity. Experimental results and theoretical analysis demonstrate that TSLoss enhances both convergence efficiency and final tuning performance. From a structural perspective, this work provides an interpretable framework for understanding soft prompt optimization and underscores the role of topological properties in guiding effective tuning. Future work will explore the relationship between structural evolution in the prompt parameter space and its semantic effects, aiming to further enhance interpretability and generalization.

\section*{Acknowledgments}
This work was supported by the National Natural Science Foundation of China (NSFC) under the General Program (Grant No. 62572104).


 

\bibliography{tda}
\bibliographystyle{IEEEtran}


 




\vfill

\clearpage

{\appendices
\section*{Supplementary Theoretical Explanation}
\label{app:1.1}
\subsection{Persistent Homology Analysis Methods}
Persistent homology \cite{edelsbrunner2002topological} is a core method in TDA \cite{wasserman2018topological} for capturing topological features such as connected components and loops in point cloud data across scales. As the distance threshold $\epsilon$ increases, the topological structure of the point cloud evolves from isolated points to connected sets and potentially higher-dimensional features such as loops. Key concepts include simplicial complexes that discretely represent point cloud topology via vertices, edges, triangles, and higher-dimensional simplices; Vietoris–Rips complexes that encode proximity relations based on a distance threshold; filtrations that track feature evolution across scales; homology groups that quantify features at each dimension with $H_0$ capturing connected components and $H_1$ capturing loops; and persistence, defined as the lifespan of features from birth to death and quantified using persistence diagrams or barcodes as well as statistics such as persistence entropy and average lifespan.

We choose to quantify the topological changes of soft prompts during training using persistent homology, with the specific analytical notation system shown in Table \ref{tab:notation}. By continuously tracking these topological features, the analysis captures persistent structures in the data that remain stable across different scales. For the analysis process of soft prompts training, the detailed step-by-step description is as follows:


\textbf{Vietoris-Rips Complex Construction.} We first obtain high-dimensional vector representations $\mathbf{x}_i$ for each soft prompt through training. To construct the topological structure of soft prompt vectors, we use the Vietoris-Rips complex, which relies on distance metrics between soft prompt vectors, specifically the Euclidean distance between vectors. Given $X = \{ \mathbf{x}_1, \mathbf{x}_2, \dots, \mathbf{x}_n \}$ as the set of soft prompt vectors, we define the complex as follows:
\begin{equation}
\begin{split}
\text{VR}_{\epsilon}(X) = \left\{ \sigma \subset X \mid \| \mathbf{x}_i - \mathbf{x}_j \| \leq \epsilon, \forall \mathbf{x}_i, \mathbf{x}_j \in \sigma \right\},
\end{split}
\end{equation}
where $\| \mathbf{x}_i - \mathbf{x}_j \|$ represents the distance between soft prompt vectors $\mathbf{x}_i$ and $\mathbf{x}_j$, and $\epsilon$ is the scale parameter indicating the maximum distance allowed for connection.

\textbf{Filtration Sequence Through Scale.} As the scale $\epsilon$ increases, we construct a sequence of nested Vietoris-Rips complexes, forming a filtration $\text{VR}_{\epsilon_1}(X) \subset \text{VR}_{\epsilon_2}(X) \subset \cdots \subset \text{VR}_{\epsilon_m}(X)$ where $\epsilon_1 < \epsilon_2 < \cdots < \epsilon_m$. This allows us to observe how the structure evolves, with richer topological features emerging as $\epsilon$ increases, such as merging of connected components and the formation and disappearance of loops.

\textbf{Homology Group Computation.} Once the Vietoris-Rips complex is constructed, we proceed to calculate its homology groups to extract topological features. We primarily focus on two types of homology groups:

The zero-dimensional homology group $H_0$ represents the connectivity of the soft prompt vector set, specifically the number of connected components in the parameter space. A lower $H_0$ value (especially at larger scales) indicates tighter connections between soft prompt vectors, resulting in a more coherent structure in the parameter space with tighter relationships between soft prompt vectors, where 
\begin{equation}
\begin{split}
H_0(X) = \frac{\text{ker} \, \partial_0}{\text{im} \, \partial_1}.
\end{split}
\end{equation}

The one-dimensional homology group $H_1$ represents the number of loops (i.e., redundancy) in the soft prompt vector set, reflecting potential structural cycles or repetitions in the soft prompt configurations. A higher $H_1$ value may indicate redundant or repetitive patterns in the soft prompt representations, increasing the structural complexity and redundancy in the soft prompt representations, where
\begin{equation}
\begin{split}
H_1(X) = \frac{\text{ker} \, \partial_1}{\text{im} \, \partial_2}.
\end{split}
\end{equation}

By analyzing how homology groups evolve across scales through persistent homology, the stability of structural properties in soft prompts, such as connectivity and redundancy, is quantitatively characterized. Both $H_0$ and $H_1$ are non-negative integers, with $H_0 \geq 1$ representing connected components and $H_1 \geq 0$ counting loops in the structure.

\textbf{Persistence Analysis} visualizes and evaluates the persistence of topological features in soft prompts. Persistence diagrams plot features as points $(b,d)$ representing birth and death scales. Points farther from the diagonal indicate more persistent features. Barcodes visualize feature lifespans as intervals $[birth, death]$. Longer bars indicate stable structures, while shorter bars may represent transient or redundant elements. Persistent Entropy quantifies the complexity in feature lifespan distribution:
\begin{equation}
\begin{split}
PE = -\sum_i \frac{l_i}{L} \log\left(\frac{l_i}{L}\right),
\end{split}
\end{equation}
where $l_i = death_i - birth_i$ is each feature's lifespan and $L = \sum_i l_i$ is the total lifespan. Lower values suggest focused, stable structures; higher values indicate more random structures with potential redundancies. Feature Lifespan Statistics include Maximum Lifespan, $\max(death_i - birth_i)$, representing the most persistent structure's stability, and Average Lifespan, $(1/n) \cdot \sum_i (death_i - birth_i)$, reflecting overall structural stability of soft prompt tuning.

\subsection{Connection Between 0-dimensional Homology Group Variance and Lifetime Distribution}
The mathematical connection between $H_0$  variance and the 0-dimensional homology group lifetime distribution $l_i$ is based on a key correspondence: the soft nearest neighbor distance $s_i$ is a differentiable approximation of the death scale $d_i$ of point $x_i$ in persistent homology. Since the lifetime of 0-dimensional features equals their death scale (the birth scale is 0), therefore:
\begin{equation}
\begin{split}
\text{Var}(\{s_i\}) \approx \text{Var}(\{d_i\}) = \text{Var}(\{l_i\})
\end{split}
\end{equation}

The expression $s_i = -\tau \log\sum_j \exp(-D_{ij}/\tau)$ provides an approximation of the nearest-neighbor distance from point $x_i$. This value determines the scale at which the point merges with other connected components in persistent homology filtration, i.e., the death scale $d_i$, thus validating the analogy.

\subsection{Persistent Entropy}
Persistent entropy is an information-theoretic measure in TDA that quantifies the uniformity of persistent homology barcode distributions. For a set of persistent homology barcodes $\{(b_i, d_i)\}_{i=1}^n$, where $b_i$ and $d_i$ denote the birth and death scales, respectively, persistent entropy is defined as
\begin{equation}
E = -\sum_{i=1}^n p_i \log(p_i),
\end{equation}
where $p_i = \frac{L_i}{L_{\text{total}}}$, $L_i = d_i - b_i$, and $L_{\text{total}} = \sum_{j=1}^n L_j$. Maximizing persistent entropy in deep learning encourages a uniform distribution of topological features and mitigates concentration in a few modes. Accordingly, lower persistent entropy typically reflects decreased structural diversity. In the context of soft prompt training, which is task-specific optimization, the learned structures tend toward simplicity and stability while preserving task-relevant semantic information. As a result, persistent entropy exhibits a moderate reduction, indicating a decrease in redundant $H_1$ features, whereas the small magnitude of change suggests that essential topological and semantic information is retained.

\section*{Additional Visualization Analysis}
\label{app2.1}
In this subsection, we present samples from additional datasets and models to illustrate the generality of our analysis.

\subsection{Additional Experimental Setups}
\label{app2.2}
The figures presented in this supplementary material are all derived from experiments conducted under the same settings as those in the main paper. They illustrate the structural evolution of soft prompt training across various language models and datasets. We randomly select and show a subset of these figures in the main paper, while the remaining images are provided in full here. All figures are generated based on structural samples collected during the training process, with experimental methods identical to those described in the main paper. The figures are organized by task type (GSM8K, MATH \cite{hendrycks2021measuring}, BBH \cite{suzgun2022challenging}, MMLU-CF \cite{zhao2024mmlu}, HotpotQA \cite{yang2018hotpotqa}, LongBench \cite{yang2018hotpotqa}) and training paradigm.

\subsection{Persistent Homology Analysis}

We conducted a unified averaging analysis of the persistent homology features, specifically $H_0$ and $H_1$, generated during both single-sample and multi-sample training processes. By comparing the persistent homology barcodes under different training settings, we observed notable intrinsic regularities in the evolution of the structures, particularly in terms of stability and consistency. These findings deepen our understanding of soft prompt tuning dynamics in the parameter space from a topological perspective, further validating the core conclusions presented in the main text regarding structural convergence and generalization capability. The following figures present the visualization results.

Figures \ref{fig:gsm-s} and \ref{fig:gsm-m} present the visualization results on the GSM8K dataset, with Figure \ref{fig:gsm-s} showing single-sample results and Figure \ref{fig:gsm-m} showing multi-sample results. Similarly, Figures \ref{fig:math-s}, \ref{fig:bbh-s}, \ref{fig:mmlu-s}, \ref{fig:hot-s}, and \ref{fig:long-s} present the single-sample visualization results on the MATH, BBH, MMLU-CF, HotpotQA, and LongBench datasets, respectively. All analyses are conducted on Gemma-2B-IT.

\begin{figure}[h!]
    \centering
    \includegraphics[width=\linewidth]{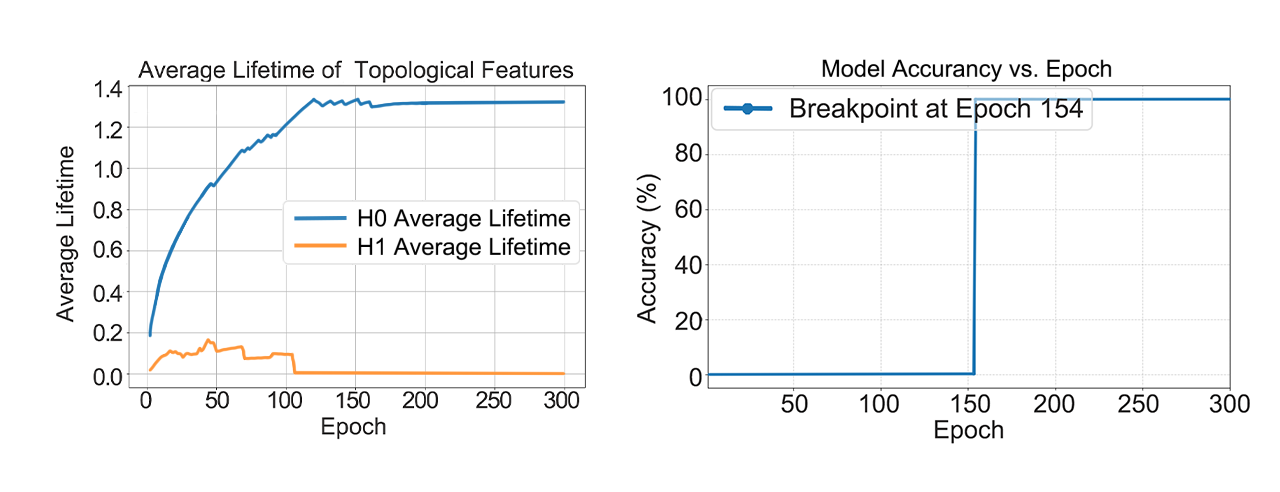}
    \caption{Single-sample results on GSM8K dataset.}
    \label{fig:gsm-s}
\end{figure}
\begin{figure}[h!]
    \centering
    \includegraphics[width=\linewidth]{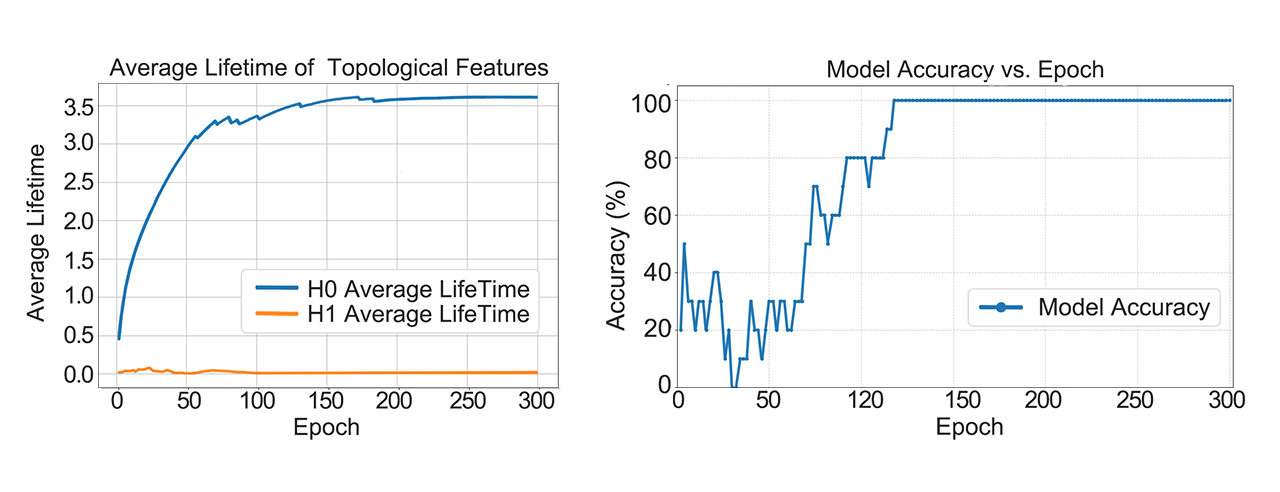}
    \caption{Multi-sample results on GSM8K dataset.}
    \label{fig:gsm-m}
\end{figure}
\begin{figure}[h!]
    \centering
    \includegraphics[width=\linewidth]{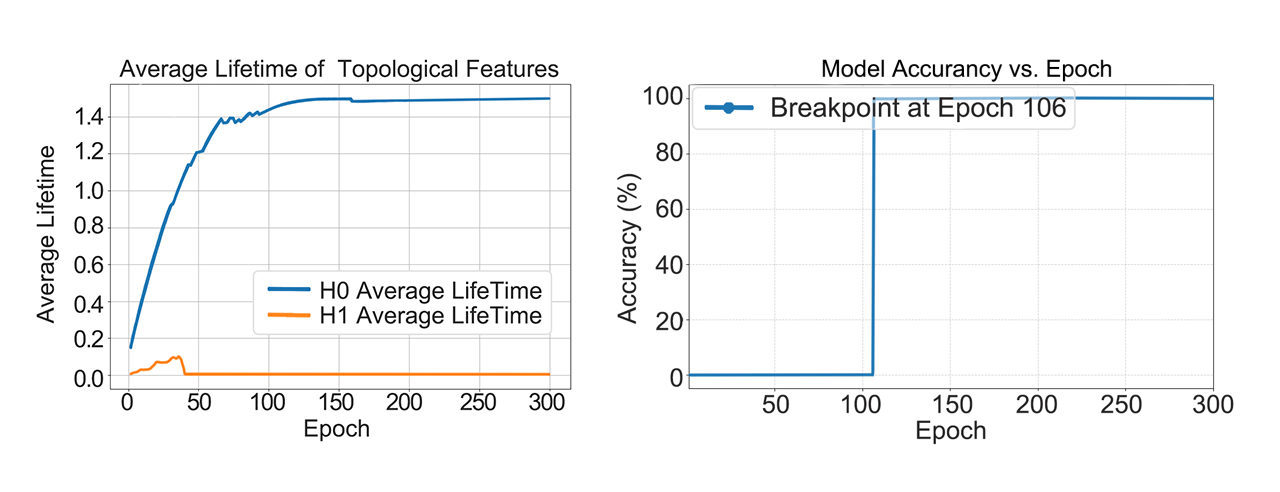}
    \caption{Single-sample results on MATH dataset.}
    \label{fig:math-s}
\end{figure}
\begin{figure}[h!]
    \centering
    \includegraphics[width=\linewidth]{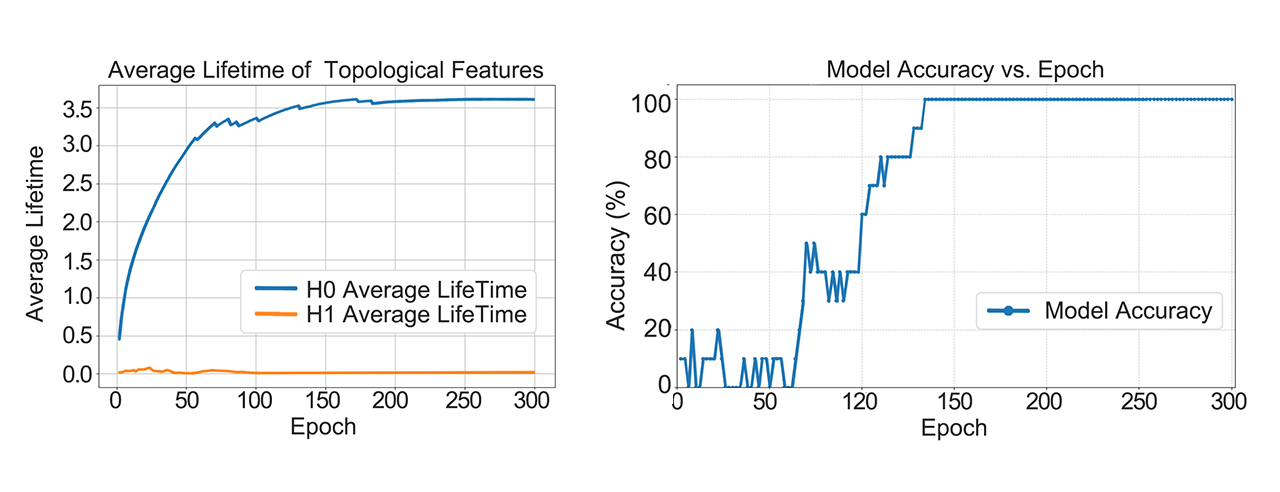}
    \caption{Multi-sample results on MATH dataset.}
    \label{fig:math-m}
\end{figure}
\begin{figure}[h!]
    \centering
    \includegraphics[width=\linewidth]{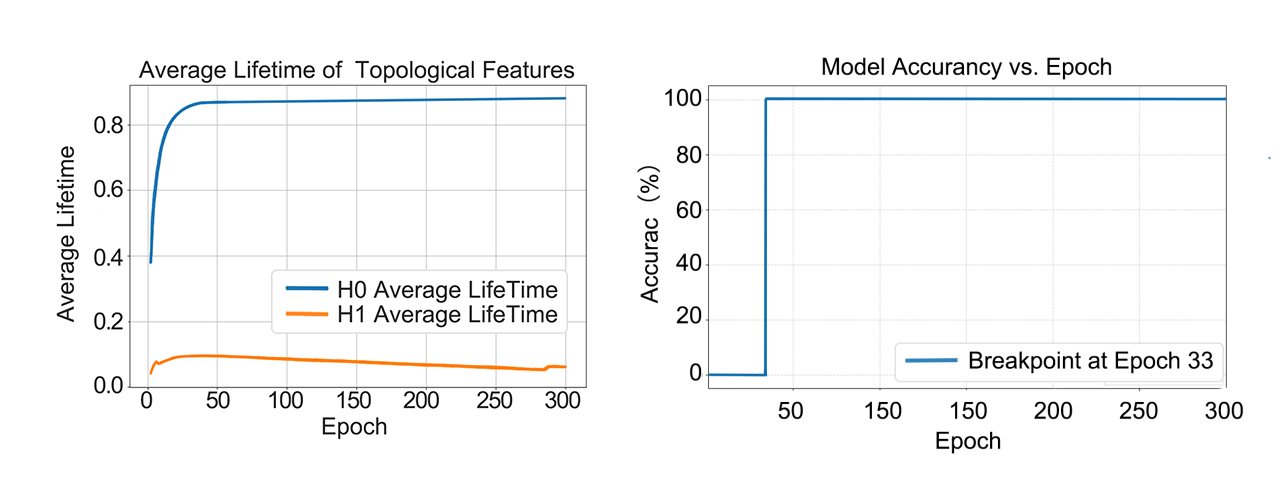}
    \caption{Single-sample results on BBH dataset.}
    \label{fig:bbh-s}
\end{figure}
\begin{figure}[h!]
    \centering
    \includegraphics[width=\linewidth]{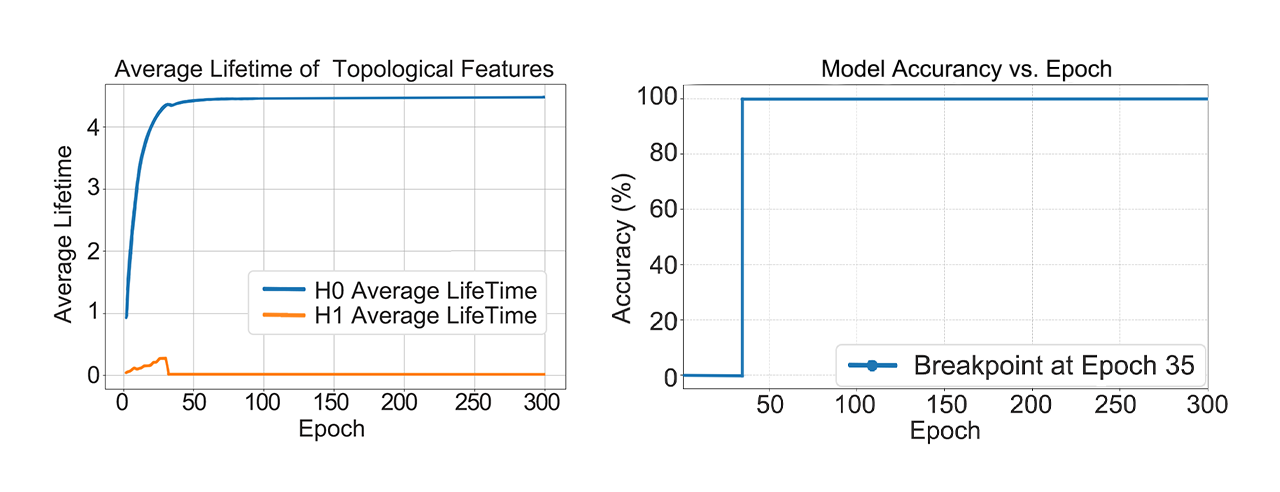}
    \caption{Single-sample results on MMLU-CF dataset.}
    \label{fig:mmlu-s}
\end{figure}
\begin{figure}[h!]
    \centering
    \includegraphics[width=0.95\linewidth]{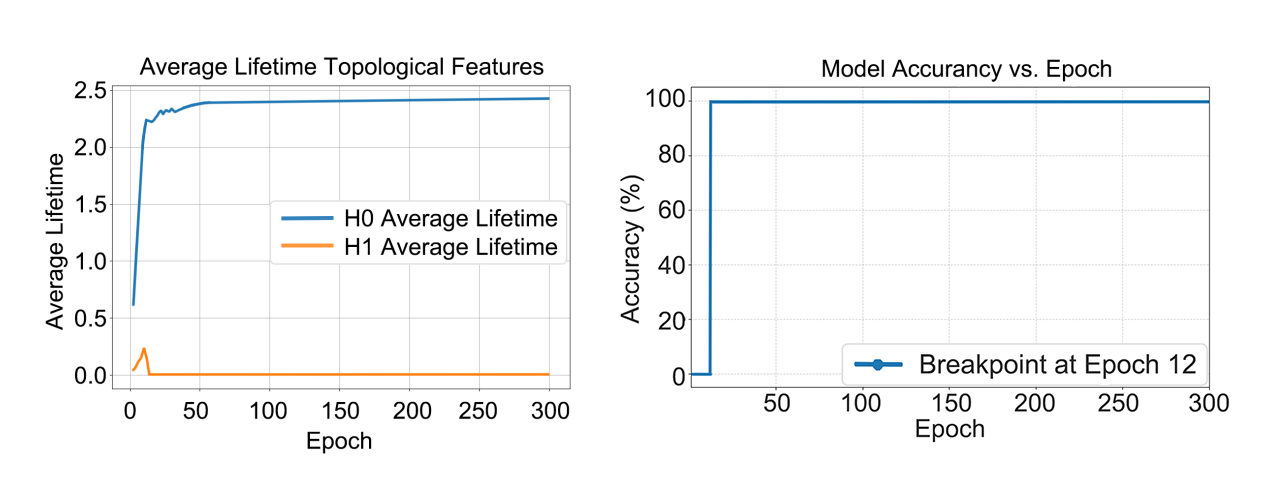}
    \caption{Single-sample results on HotpotQA dataset.}
    \label{fig:hot-s}
\end{figure}
\begin{figure}[h!]
    \centering
    \includegraphics[width=0.95\linewidth]{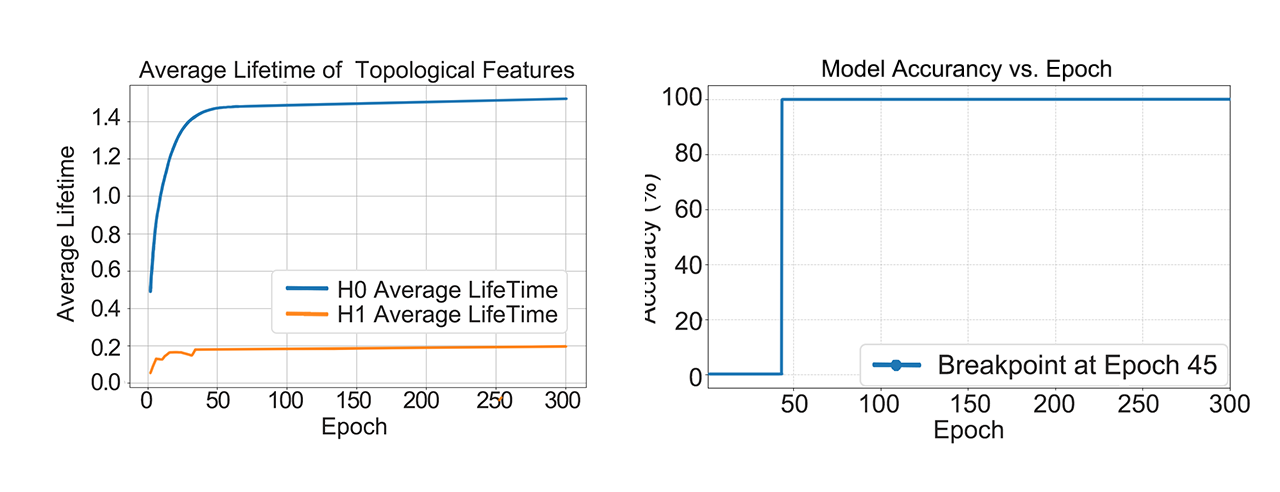}
    \caption{Single-sample results on LongBench dataset.}
    \label{fig:long-s}
\end{figure}
}

\end{document}